\documentclass[letterpaper, 10 pt, conference]{ieeeconf}  

\IEEEoverridecommandlockouts                              

\usepackage[backend=biber,
            hyperref=true,
            url=false,
            isbn=false,
            doi=false,
            backref=false,
            style=ieee,
            citestyle=numeric-comp,
            sorting=nyt,
            block=none]{biblatex}

\usepackage[font=footnotesize]{caption}

\renewcommand{\bibfont}{\small}
\usepackage{amsmath}
\usepackage{amssymb}
\usepackage{graphicx}
\usepackage{breqn}
\usepackage{subcaption}
\usepackage{xspace}
\usepackage[dvipsnames]{xcolor}
\usepackage{wrapfig}
\usepackage{relsize}

\usepackage{algorithm}
\usepackage{algcompatible}

\usepackage{hyperref}
\usepackage[capitalise]{cleveref}

\usepackage{tabularx}
\usepackage{array}

\usepackage[hang,flushmargin]{footmisc}
\usepackage{url}

\usepackage[export]{adjustbox}
\newcommand{\METHOD}{MAPLE~}
\newcommand{\METHODNOSPC}{MAPLE}
\newcommand{\METHODLONG}{\underline{Ma}nipulation \underline{P}rimitive-augmented reinforcement \underline{Le}arning}
\newcommand{\METHODLONGHIGHLIGHT}{\textbf{Ma}nipulation \textbf{P}rimitive-augmented reinforcement \textbf{Le}arning}

\newcommand{\pihi}{\pi_{tsk}}
\newcommand{\pilo}{\pi_{p}}

\newcommand{\Q}[0]{Q_\theta}
\newcommand{\Qbar}[0]{Q_{\bar{\theta}}}

\newcommand{\pihip}[0]{\pi_{tsk_\phi}}
\newcommand{\pilop}[0]{\pi_{p_\psi}}

\DeclareMathOperator*{\E}{\mathbb{E}}

\def\citep{\cite}
\def\citet{\cite}

\title{\LARGE \bf
Augmenting Reinforcement Learning with Behavior Primitives\\for Diverse Manipulation Tasks
}

\author{Soroush Nasiriany$^{1}$, Huihan Liu$^{1}$, Yuke Zhu$^{1}$
\thanks{\fontsize{7}{8}\selectfont $^{1}$The University of Texas at Austin. Correspondence: \texttt{soroush@cs.utexas.edu}}%
}

\DeclareUnicodeCharacter{0301}{\'{i}}
\begin{document}

\maketitle
\thispagestyle{empty}
\pagestyle{empty}

\begin{abstract}
    Realistic manipulation tasks require a robot to interact with an environment with a prolonged sequence of motor actions. While deep reinforcement learning methods have recently emerged as a promising paradigm for automating manipulation behaviors, they usually fall short in long-horizon tasks due to the exploration burden.
This work introduces \METHODLONG~(\METHODNOSPC), a learning framework that augments standard reinforcement learning algorithms with a pre-defined library of behavior primitives. These behavior primitives are robust functional modules specialized in achieving manipulation goals, such as grasping and pushing. To use these heterogeneous primitives, we develop a hierarchical policy that involves the primitives and instantiates their executions with input parameters.
We demonstrate that \METHOD outperforms baseline approaches by a significant margin on a suite of simulated manipulation tasks. We also quantify the compositional structure of the learned behaviors and highlight our method's ability to transfer policies to new task variants and to physical hardware.
Videos and code are available at \url{https://ut-austin-rpl.github.io/maple}
\end{abstract}

\section{Introduction}
\label{sec:intro}

Enabling autonomous robots to solve diverse and complex manipulation tasks has been a grand challenge for decades. In recent years, deep reinforcement learning (DRL) approaches have made great strides towards designing robot manipulation behaviors that are difficult to engineer manually~\cite{kalashnikov2018qtopt, openai2019learning,openai2019solving,kalashnikov2021mtopt}. Nonetheless, state-of-the-art DRL models fall short in long-horizon tasks due to the exploration challenge --- the robot has to explore a prohibitively large space of possible behaviors for accomplishing a task.
To remedy the exploration burden, prior DRL work has developed various temporal abstraction frameworks to exploit the hierarchical structure of manipulation tasks~\cite{coreyes18sectar,nachum2018hiro,bacon2016optioncritic,eysenbach2019sorb}. These methods learn low-level controllers, often modeled as \textit{skills} or \textit{options}, together with high-level controllers from trial-and-error. While they have demonstrated greater scalability than vanilla DRL methods, they often suffer from high sample complexity, lack of interpretability, and brittle generalization.

In the meantime, decades-long research in robotics has developed a rich repertoire of functional modules specialized at particular robot behaviors, such as grasping~\cite{bohg2014graspsurvey} and motion planning~\cite{karaman2011rrtstar,ijspeert2013dmps}. These pre-built functional modules, which we refer to as \emph{behavior primitives}, exhibit a high degree of robustness and reusability for achieving certain manipulation goals, such as picking up objects with the end-effector and moving the robot to a target configuration in a collision-free path. In spite of their specialties, it remains a challenge for DRL algorithms to use them as the building blocks to scaffold complex tasks. The challenge is primarily due to the fact that these behavior primitives are \textit{heterogeneous} by design. They take non-uniform parameters as input, operate at varying temporal resolutions, and exhibit distinct behaviors. This thus requires an algorithm to reason about the temporal decomposition of a complex task and adaptively compose these behavior primitives accordingly.

A variety of hierarchical modeling approaches in robotics have used behavior modules as low-level building blocks. Notably, task-and-motion planning~\cite{kaelbling2011hpn, garrett2020tamp, wang2021comptamp} and neural programming~\cite{xu2018neural, huang2019neural} methods have used primitives such as motion planners and pick-and-place controllers to model manipulation tasks in a compositional fashion. They require well-specified domain knowledge to perform task planning or strong human supervision to train a high-level controller with ground-truth task decomposition. These assumptions limit the scalability of these methods in realistic tasks.

\begin{figure}[t]
    \centering
    \includegraphics[width=1.0\linewidth]{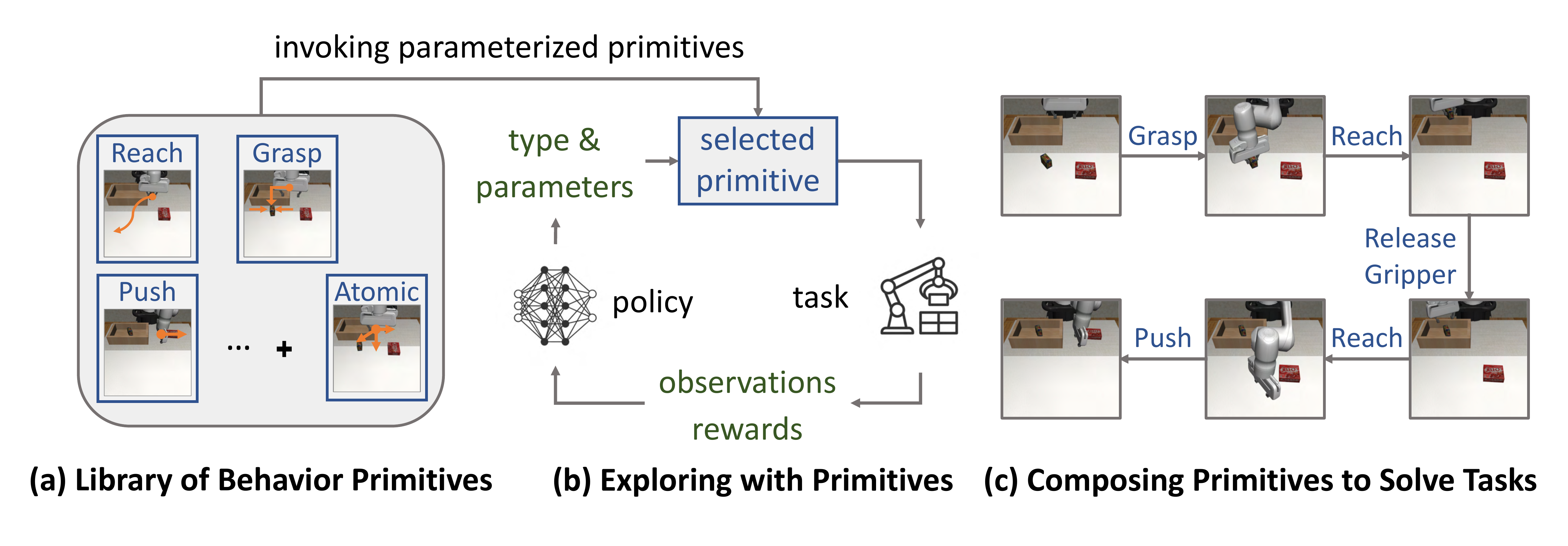}
    \caption{{\bf Overview of \METHODNOSPC.} (a) We present a learning framework that augments the robot's atomic motor actions with a library of versatile behavior primitives. (b) Our method learns to compose these primitives via reinforcement learning. (c) This enables the agent to solve complex long-horizon manipulation tasks.}
    \label{fig:overview}
\end{figure}

In this work, we introduce \METHODNOSPC~(\METHODLONGHIGHLIGHT), a general DRL algorithm that harnesses a set of pre-built behavior primitives for solving long-horizon manipulation tasks.
To address the exploration challenge of DRL algorithms, our method uses a library of high-level behavior primitives (such as grasping or pushing objects) in conjunction with low-level motor actions to autonomously learn a hierarchical policy (see \cref{fig:overview}).
Our algorithm models each behavior primitive as an implementation-agnostic controller that produces a temporally extended behavior.
At a given state, our DRL policy invokes a behavior primitive (or an atomic motor action) and instantiates it with input parameters. For example, the input parameters to a 6-DoF grasping module can be the pre-grasp end-effector pose. The selected primitive interprets the input parameters and executes one or a sequence of motor actions to realize its specialized behavior.
By integrating behavior primitives into DRL algorithms, \METHODNOSPC~shields away a substantial portion of complexity in manipulation planning, while leaving the flexibility to a generic reinforcement learning algorithm to discover the compositional structure of tasks without strong domain knowledge.
Furthermore, by retaining low-level motor actions \METHODNOSPC~can rely on these actions for the stages of tasks where the finite library of behavior primitives is insufficient to express a desired behavior.

We conduct an extensive set of experiments on a suite of eight manipulation tasks of varying complexities in the robosuite simulation framework~\cite{robosuite2020}. We compare our method to standard DRL approaches~\cite{haarnoja2018sac} that only use low-level motor actions,  hierarchical DRL methods that learn options~\cite{zhang2019dac,nachum2018hiro,chitnis2020schema} or open-loop task schemas~\cite{chitnis2020schema}. \METHODNOSPC~achieves a $70\%$ increase in task success rate compared to using only atomic actions, becoming the only method that consistently solved all single-arm tasks in the standard robosuite benchmark.
We also devise a data-driven metric to quantitatively examine the compositionality of manipulation tasks contingent on the available primitives, offering new insight on the challenges and opportunities of compositional modeling for realistic manipulation tasks.

We highlight three contributions of this work: 1) We develop a novel method that augments standard DRL algorithms with pre-defined behavior primitives to reduce the exploration burden; 2) We validate the effectiveness of our method in solving diverse manipulation tasks and quantitatively analyze the compositional structure of these tasks; and 3) We show that the modularity and abstraction offered by the behavior primitives facilitate knowledge transfer of the learned policies to new task variants and physical hardware.	
\section{Related Work}
\label{sec:related}

\noindent \textbf{Deep Reinforcement Learning.}
Prior work on DRL has investigated a number of approaches to solve long-horizon tasks, through improved exploration strategies~\cite{bellemare2016unifying,pathak2017curiositydriven,houthooft2017vime,pathak2019selfsupervised}, learning options~\cite{nachum2018hiro,bacon2016optioncritic,smith2018inference,zhang2019dac,bagaria2020dsc}, unsupervised skill discovery~\cite{eysenbach2018diversity,sharma2020dynamicsaware}, and integrating planning~\cite{eysenbach2019sorb,nasiriany2019planning}.
Despite these efforts, today's DRL methods still struggle in long-horizon robotic tasks due to the exploration burden of learning from scratch. Recent work has examined the use of offline data to alleviate the exploration burden in DRL, through demonstration-guided RL~\cite{rajeswaran2018learning, nair2018overcoming,gupta2019relay}, learned behavioral priors~\cite{pertsch2020spirl, singh2020parrot} and action spaces~\cite{ajay2020opal,allshire2021laser} from demonstrations, and offline RL~\cite{fujimoto2019offpolicy,mandlekar2020iris,kumar2020conservative,fu2021d4rl}.
While promising, these methods can be difficult to scale up due to the costs of acquiring offline data.

\noindent \textbf{Hierarchical Modeling in Robotics.}
Outside of DRL, there has been a plethora of work in robotics dedicated to building customized functional modules that emit specific robot behaviors, such as grasping~\cite{bohg2014graspsurvey,mahler2017dexnet2} and motion planning~\cite{karaman2011rrtstar,amato1996randomized}. Prior works on task-and-motion planning~\cite{kaelbling2011hpn, garrett2020tamp, wang2021comptamp} and neural programming~\cite{xu2018neural, huang2019neural} have developed hierarchical models that leverage these modules as building blocks to scaffold manipulation tasks. While these methods have demonstrated impressive capabilities in restrictive domains, their applicability has been limited by their reliance on domain knowledge or human supervision.

To bridge the gap between these models and DRL algorithms that learn from scratch, recent work has harnessed pre-built primitives, such as model-based planners~\cite{lee2020guapo}, motion planners~\cite{yamada2020mopa,xiali2020relmogen}, movement primitives~\cite{ijspeert2013dmps,neumann2014skills}, and pre-built skills~\cite{chitnis2020schema,lee2020skills, strudel2020skills,sharma2020skills,simeonov2020skills}, to expedite DRL algorithms.
These approaches aim at retaining the flexibilities of RL algorithms while benefiting from the temporal abstraction provided by the primitives.
However, these works are limited as they are confined to using only one or two specific primitives~\cite{lee2020guapo,yamada2020mopa,xiali2020relmogen}, employ rigid primitives that are not reconfigurable~\cite{strudel2020skills,sharma2020skills}, or hard-code how the primitives are composed~\cite{simeonov2020skills}.
In contrast, our method adopts a set of versatile primitives to solve diverse manipulation tasks.

\noindent \textbf{Reinforcement Learning with PAMDPs.} Our formalism specifically falls under the established reinforcement learning framework of Parameterized Action MDPs (PAMDPs)~\cite{masson2015reinforcement}, in which the agent executes a parameterized primitive at each decision-making step.
We note that several prior works~\cite{hausknecht2016deep, wei2018hierarchical, xiong2018parametrized, fan2019hybrid, jain2020actions} have adapted off-the-shelf deep RL algorithms to the PAMDP setting. Nonetheless, they have focused on relatively simple game domains, shielding away practical challenges in robot manipulation, such as high-dimensional continuous state/action spaces and heterogeneous primitives.
We note that Chitnis et al.~\citet{chitnis2020schema} and Lee et al.~\citet{lee2020skills} have modeled robot manipulation with PAMDPs. We provide empirical comparisons to demonstrate the limitations of their modeling choices, yielding less competitive performance in challenging manipulation tasks than ours.
Concurrent work by Dalal et al.~\citep{dalal2021raps} also studies the application of robotic primitives for manipulation tasks.
Our work complements theirs with additional analysis on the compositional structure of the learned behavior and experiments demonstrating the ability to transfer learned policies to novel task variants and to physical hardware.
\section{Method}
\label{sec:prelim}

Our goal is to enable robots to leverage behavior primitives to solve manipulation tasks effectively and efficiently.
To that end, we seek a library of behavior primitives that serve as the building blocks to scaffold manipulation tasks and a reinforcement learning algorithm that composes these primitives to solve tasks.
To evaluate whether our algorithm facilitates compositional behaviors, we also propose a metric to quantify the degree to which the resulting learned behavior is compositional.
See \cref{fig:overview} for an overview of our method.

\subsection{Decision-Making with Parameterized Behavior Primitives}
\label{sec:framework}
We adopt reinforcement learning (RL) as the underlying decision-making framework.
The objective of RL is to maximize the expected infinite sum of discounted rewards in a Markov Decision Process (MDP), defined by the tuple $\mathcal{M} = (\mathcal{S}, \mathcal{A}, r, p, p_{0}, \gamma )$.
The entities in the tuple represent the state space, the action space, the reward function, the transition function, the initial state distribution, and the discount factor.
In most robotic RL problems, the action space $\mathcal{A}$ consists of all atomic actions $u \in \mathbb{R}^{d_{\text{control}}}$ provided by the robot, such as end-effector displacements.
We augment this action space with a heterogeneous library of behavior primitives $\mathcal{L} = \{a^1, a^2, \cdots, a^k \}$ that perform semantically meaningful behaviors.
Formally, each behavior primitive $a \in \mathcal{L}$ is represented by a control module $M_a(x)$ that executes a finite, variable sequence of atomic actions $(u_1, u_2, \cdots, u_t), u_i \in \mathbb{R}^{d_{\text{control}}}$, where the exact action sequences are specified by input parameters $x \in \mathbb{R}^{d_a}$. Here $d_a$ is the dimension of the input parameters to the primitive $a$ that varies across different primitives. To incorporate these behavior primitives, we recast our decision-making problem as a Parameterized Action MDP (PAMDP)~\cite{masson2015reinforcement}, where at each decision-making step the agent executes a parameterized action $(a, x) \in \mathcal{A}$ consisting of the type of primitive $a$ and its parameters $x$.

\subsection{Behavior Primitives: Building Blocks for Manipulation}
\label{sec:primitives}
We are interested in equipping agents with a  library of versatile primitives that serve as the core building blocks for diverse manipulation tasks. 
Our decision-making algorithm assumes no knowledge of the implementations of these primitives; these primitives can comprise closed-loop skills learned via reinforcement ~\cite{haarnoja2018sac,schulman2017ppo} or imitation learning~\cite{osa2018il}, analytical motion planners~\cite{karaman2011rrtstar}, and even full-fledged grasping systems~\cite{mahler2017dexnet2,bohg2014graspsurvey}.
Regardless of their inner workings, we must ensure that our primitives are versatile and adaptive to behavioral variations. In our learning framework, we consider these primitives as \emph{functional APIs} that take input parameters $x$ that instantiate action execution. The input parameters usually have clear semantics, such as the 6-DoF end-effector pose for a grasping primitive or a target robot configuration for a motion planning primitive.
We recognize that our library of primitives may still not be universally applicable in every setting, and equipping the agent solely with these primitives may limit the set of possible behaviors that the agent can achieve. We address this limitation by introducing an additional \textit{atomic primitive} $a^{\text{atom}}$ dedicated to performing atomic robot actions to fill in any missing gaps that cannot be fulfilled by the other primitives.

In this work, we design a library of five primitives, including prehensile and non-prehensile motions, that forms the basis for many manipulation tasks:
\begin{itemize}
    \setlength\itemsep{0.3em}
    \item \textbf{Reaching:} The robot moves its end-effector to a target location $(x,y,z)$, specified by the input parameters. Execution takes at most 15 atomic actions. 
    \item \textbf{Grasping:} The robot moves its end-effector to a pre-grasp location $(x,y,z)$ at a yaw angle $\psi$, specified by the input parameters, and closes its gripper. Execution takes at most 20 atomic actions. 
    \item \textbf{Pushing:} The robot reaches a starting location $(x,y,z)$ at a yaw angle $\psi$ and then moves its end-effector by a displacement $(\delta_x, \delta_y, \delta_z)$. The input parameters are 7D. Execution takes at most 20 atomic actions.
    \item \textbf{Gripper Release:} The robot repeatedly applies atomic actions to open its gripper. This primitive has no input parameters.  Execution takes 4 atomic actions.
    \item \textbf{Atomic:} The robot applies an atomic action $\in \mathbb{R}^{d_{\text{control}}}$.
\end{itemize}
We implemented these primitives as hard-coded closed-loop controllers, each requiring only a handful of lines of code.
We highlight that these primitives take input parameters of different dimensions, operate at variable temporal lengths, and produce distinct behaviors. These properties make them challenging to utilize in a learning framework. In the following, we will introduce our algorithm for composing these primitives to solve diverse manipulation tasks.

\begin{figure}
    \centering
    \includegraphics[width=1.0\linewidth]{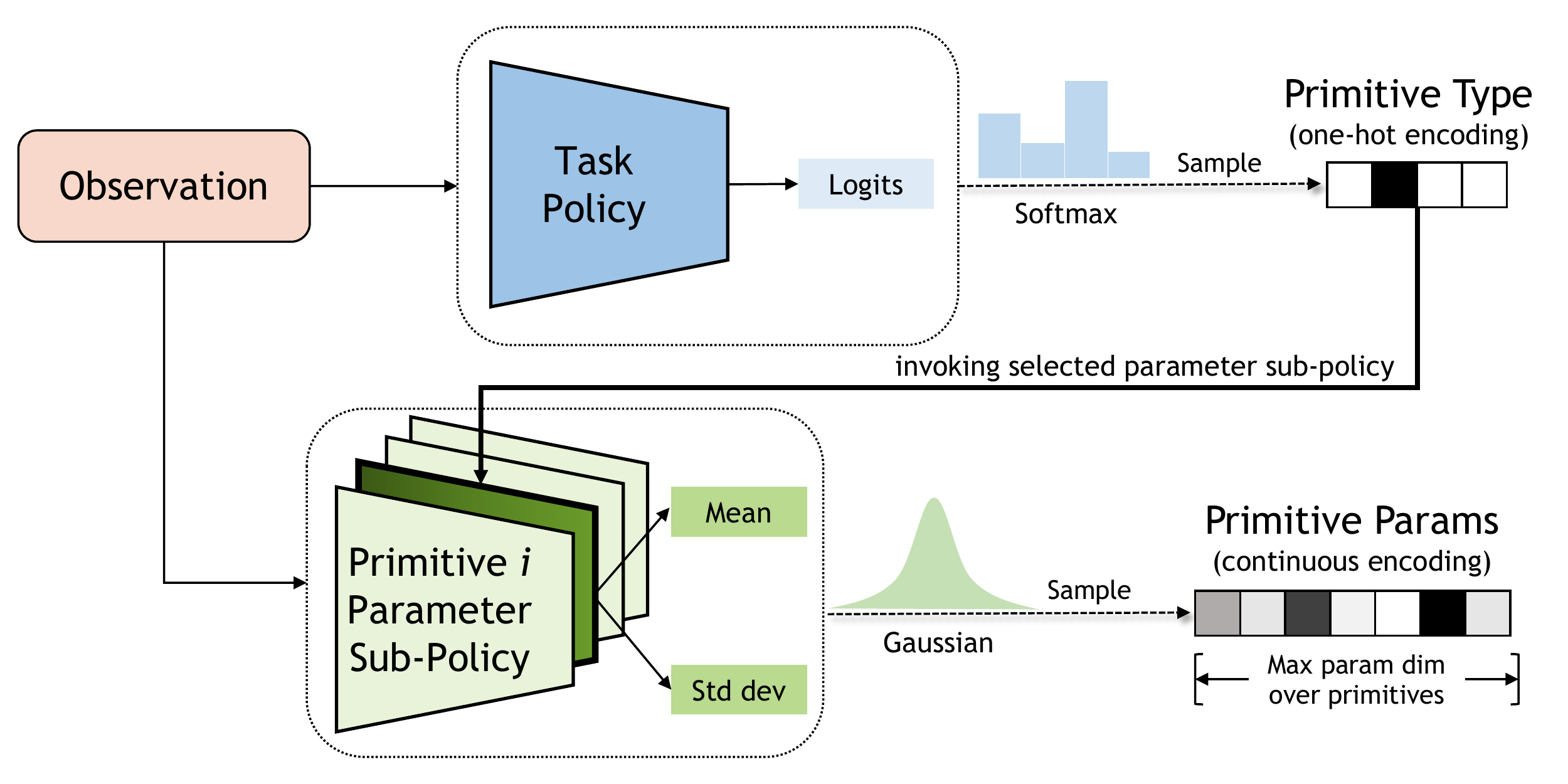}
    \caption{\textbf{Policy Architecture.}
    We adopt a hierarchical policy, with a high-level task policy that determines \textit{which} primitive to apply and a low-level parameter policy that determines \textit{how} to instantiate that primitive.
    }
    \label{fig:policy-architecture}
\end{figure}

\subsection{Composing Primitives via Reinforcement Learning}
\label{sec:algo}
We follow the PAMDP framework outlined in \cref{sec:framework}.
Previous work has explored various policy structures that reason over parameterized primitives.
The simplest approach is a flat policy~\cite{xiali2020relmogen,hausknecht2016deep} that outputs a distribution over the primitive type $a$ and all primitive parameters $\{x^1, x^2, \cdots, x^k\}$.
A major drawback of this approach is that the total number of policy outputs can quickly become intractable as additional primitives are introduced.
We address this limitation with a hierarchical policy where at the high level a \textit{task policy} $\pihi$ determines the primitive type $a$ and at the low level a \textit{parameter policy} $\pilo$ determines the corresponding primitive parameters $x$.
For implementation, we represent the task policy as a single neural network and the parameter policy as a collection of sub-networks, with one sub-network dedicated for each primitive.
This enables us to accommodate primitives with heterogeneous parameterizations.
To allow batch tensor computations across primitives with different parameter dimensions, these parameter policy sub-networks all output a ``one size fits all'' distribution over parameters $x \in \mathbb{R}^{d_A}$, where $d_A = \max_a d_a$ is the maximum parameter dimension over all primitives.
At primitive execution we simply truncate the parameters $x$ to the length $d_a$ of the chosen primitive $a$.
See \cref{fig:policy-architecture} for an illustration of our policy architecture.
In addition to reducing the overall number of output parameters our hierarchical design facilitates modular reasoning, delegating the high-level to focus on \textit{which} primitive to execute and the low-level to focus on \textit{how} to instantiate that primitive.
We note that a few prior works have previously explored this hierarchical design~\cite{wei2018hierarchical,fan2019hybrid} but to our knowledge we are the first to demonstrate its utility on complex manipulation domains with a set of heterogeneous primitives.

In principle, we can  integrate our policy architecture with any DRL algorithm designed for continuous control; we choose Soft Actor-Critic (SAC)~\cite{haarnoja2018sac}.
We modify the standard critic neural network $ \Q(s, a)$ and actor neural network $ \pi_{\phi}(a|s)$ with our critic network $\Q(s, a, x)$ and our hierarchical policy networks $\pihip(a | s)$, $\pilop(x | s, a)$.
The losses for the critic, task policy, and parameter policy are defined respectively (with components pertaining to the \textcolor{BrickRed}{task policy in red} and the \textcolor{RoyalBlue}{parameter policy in blue}):
\resizebox{\linewidth}{!}{
\begin{minipage}{\linewidth}
\begin{align*}
J_Q(\theta) {} &= \bigg( \Q(s, a, x) - \Big( r(s, a, x) + \gamma \Big(\Qbar(s', a', x')
\\ 
&- \textcolor{BrickRed}{\alpha_{tsk} \log(\pihip(a' | s'))} - \textcolor{RoyalBlue}{\alpha_p \log(\pilop(x' | s', a'))} \Big) \Big) \bigg)^2 \notag &
\\[10pt]
J_{\pihi}(\phi) &= \E_{a \sim \pihip} \Big[ \textcolor{BrickRed}{\alpha_{tsk} \log(\pihip(a | s))} - \E_{x \sim \pilop} \Q(s, a, x) \Big]
\\[10pt]
J_{\pilo}(\psi) &= \E_{a \sim \pihip} \E_{x \sim \pilop} \Big[ \textcolor{RoyalBlue}{\alpha_{p} \log(\pilop(x | s, a))} - \Q(s, a, x) \Big]
\\
\end{align*}
\end{minipage}
}
Here $\alpha_{tsk}$ and $\alpha_{p}$ control the maximum entropy objective for the task policy and parameter policy, respectively.

\subsection{Facilitating Exploration with Affordances}
\label{sec:affordance}
Compared with existing methods that reason purely over atomic actions, our algorithm benefits from accelerated exploration due to the temporal abstraction provided by our behavior primitives.
However, even reasoning with temporally extended actions can present an exploration challenge~\cite{pertsch2020spirl}.
One way to address this issue is to equip the agent with \textit{affordances} that help to discern the utility of actions in different settings.
For example, a grasping skill is only appropriate when applied in the vicinity of graspable objects, and a pushing skill is only appropriate in the vicinity of pushable objects.
In our framework, these affordances can be expressed by adding to the reward function an auxiliary affordance score $s_{\text{aff}}(s, x; a) \in [0,1]$ that measures the affinity for parameters $x$ at a particular state $s$ for a given primitive $a$.
These affordances scores can in principle come from learned models trained on robot interaction data~\cite{simeonov2020skills,nagarajan2020learning,mandikal2020graff,mo2021where2act,xu2021daf} or human data~\cite{do2018affordance,fang2018demo2vec,nagarajan2019grounded,kokic2020learning}.
Nonetheless, as our primitive parameters carry clear semantic meanings, we can analytically define these affordance scores based on the objects' physical states.
Concretely, for the \textit{atomic} and \textit{gripper release} primitives, we always give an affordance score of $1$ to enable the universal applicability of these primitives.
For the remaining \textit{reach}, \textit{grasp}, and \textit{push} primitives we implement general, easy-to-define affordances encouraging the agent to reach relevant areas of interest in the workspace.
More specifically, these primitives all involve reaching a location $x_{reach} = \texttt{[x[0], x[1], x[2]]}$ and we encourage the agent to specify the reaching parameters $x_{reach}$ to be within a threshold of a set of keypoints $P$:
\resizebox{\linewidth}{!}{
\begin{minipage}{\linewidth}
\begin{align*}
s_{\text{aff}}(s, x; a) &= \max_{p \in P} \ 1 - \tanh\Big(\max(\Vert x_{reach} - p \Vert - \tau, 0)\Big)
\end{align*}
\vspace{1pt}
\end{minipage}
}
The keypoints $P$ are the locations of objects to push for the push primitive, the locations of objects to grasp for the grasp primitive, and the locations of reaching targets for the reach primitive.

\begin{figure}
    \setlength{\belowcaptionskip}{0pt}
    \vspace{2mm}
    \centering
    \begin{subfigure}[t]{0.24\linewidth}
        \includegraphics[width=\textwidth]{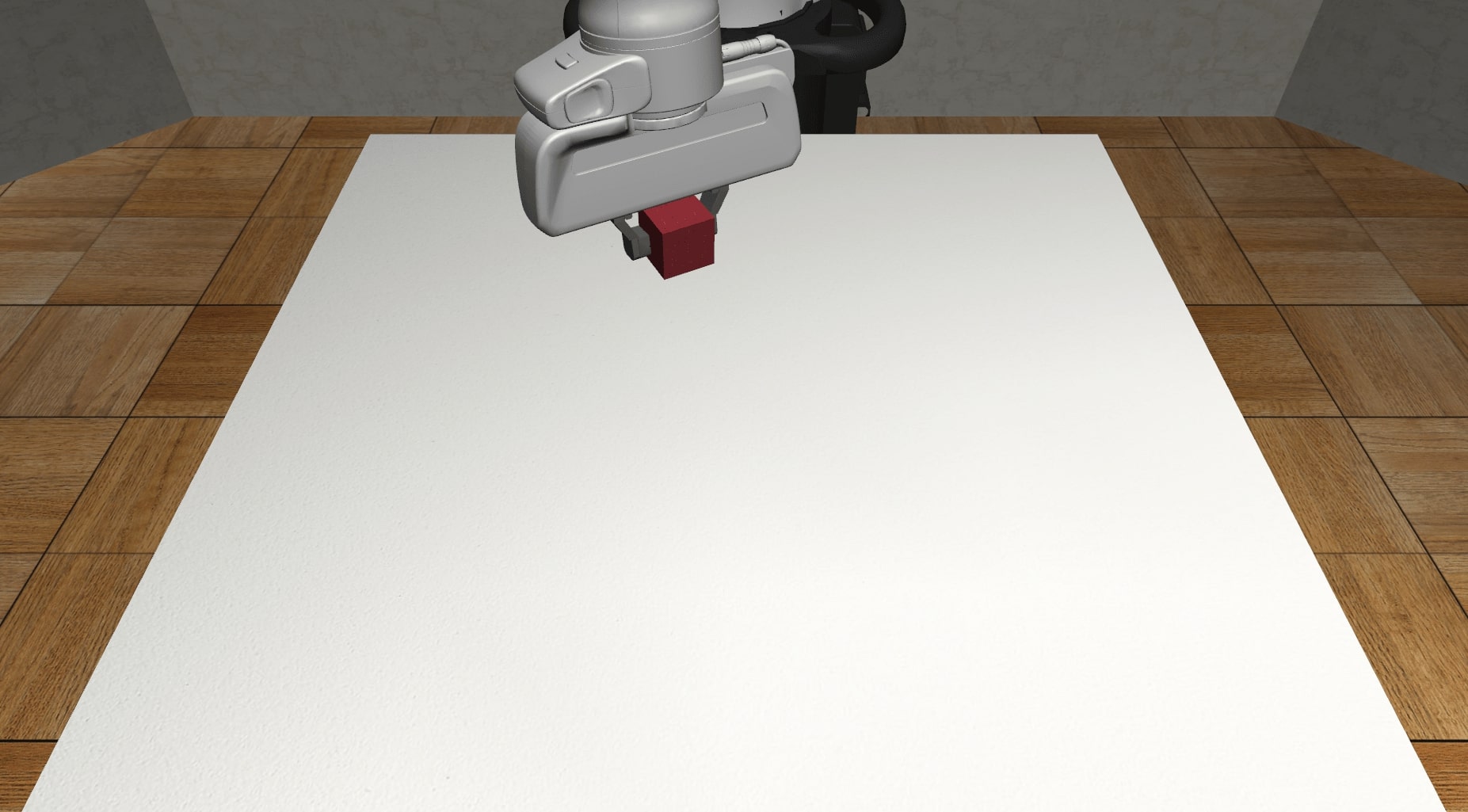}
        \vspace{-6mm}
        \caption*{\textbf{Lift}}
    \end{subfigure}
    \hfill
    \begin{subfigure}[t]{0.24\linewidth}
        \includegraphics[width=\textwidth]{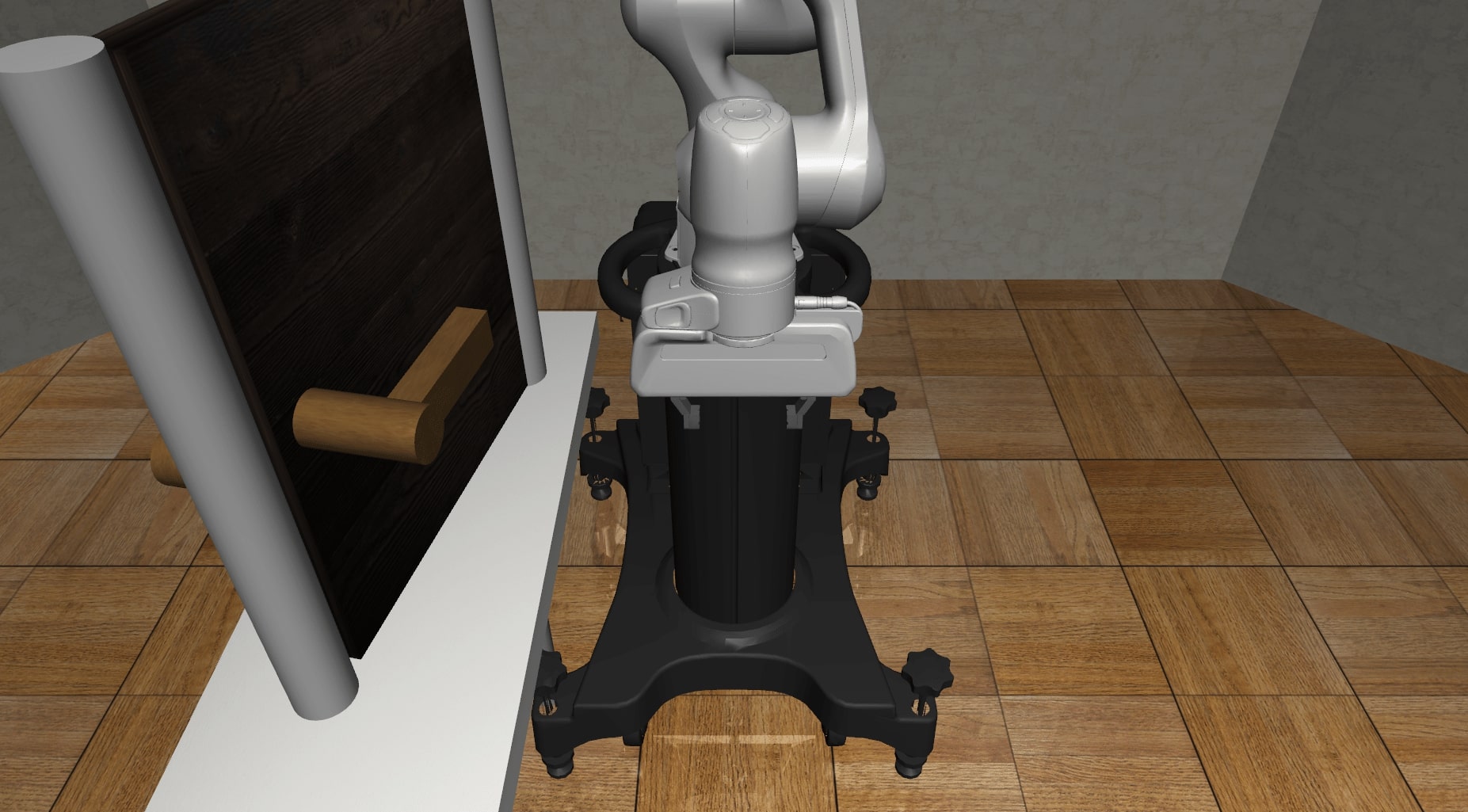}
        \vspace{-6mm}
        \caption*{\textbf{Door Opening}}
    \end{subfigure}
    \hfill
    \begin{subfigure}[t]{0.24\linewidth}
        \includegraphics[width=\textwidth]{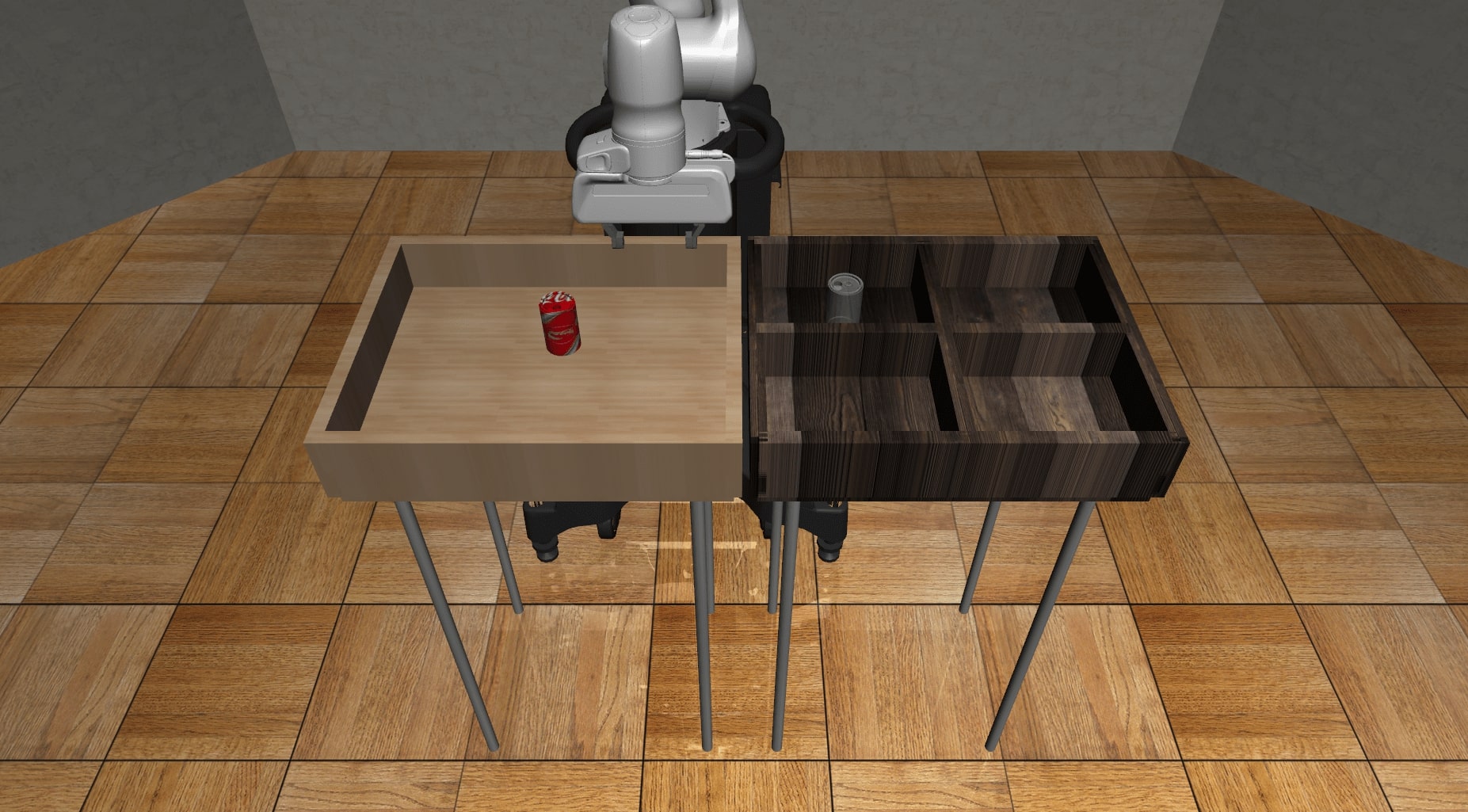}
        \vspace{-6mm}
        \caption*{\textbf{Pick and Place}}
    \end{subfigure}
    \hfill
    \begin{subfigure}[t]{0.24\linewidth}
        \includegraphics[width=\textwidth]{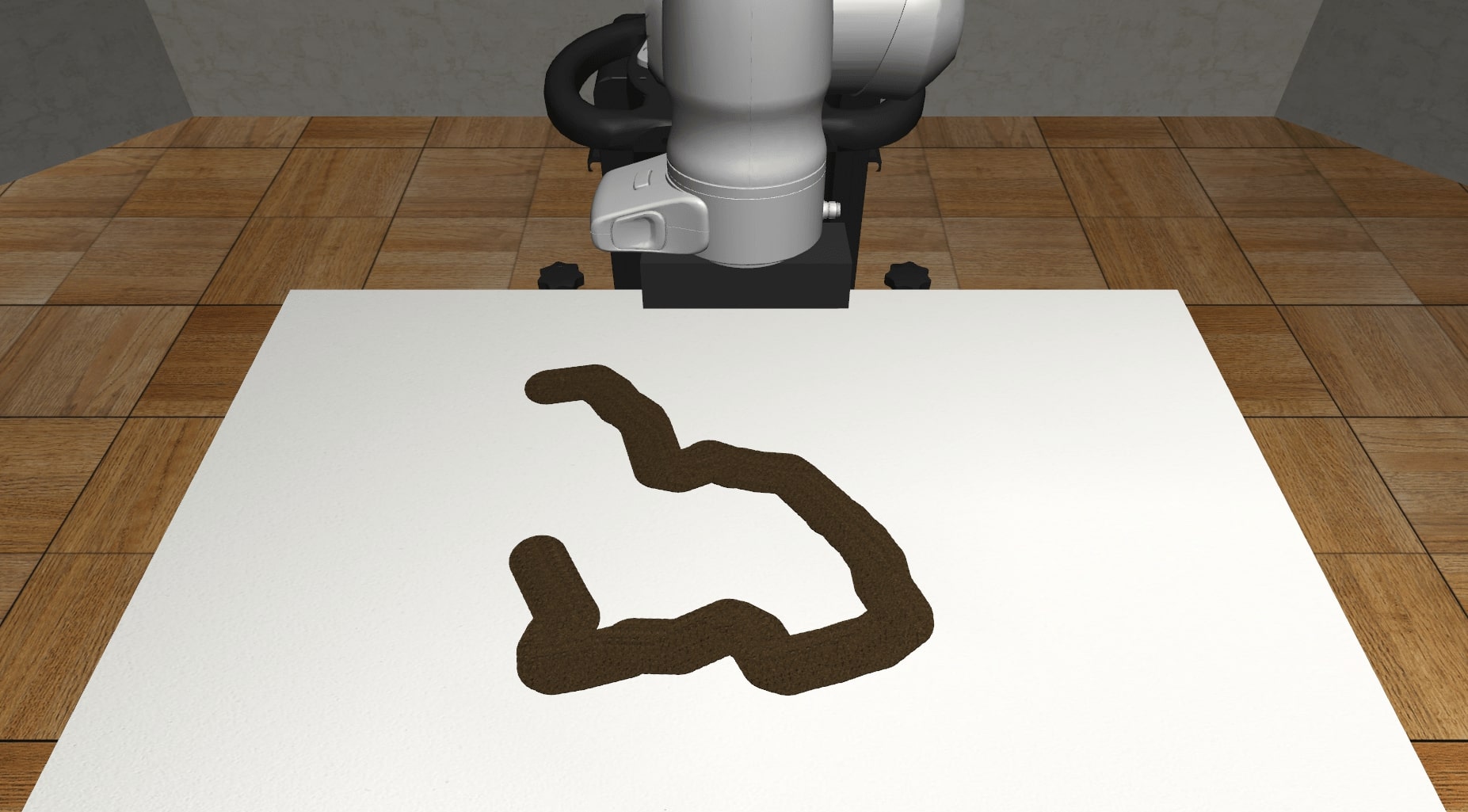}
        \vspace{-6mm}
        \caption*{\textbf{Wipe}}
    \end{subfigure}
    \hfill
    \begin{subfigure}[t]{0.24\linewidth}
        \includegraphics[width=\textwidth]{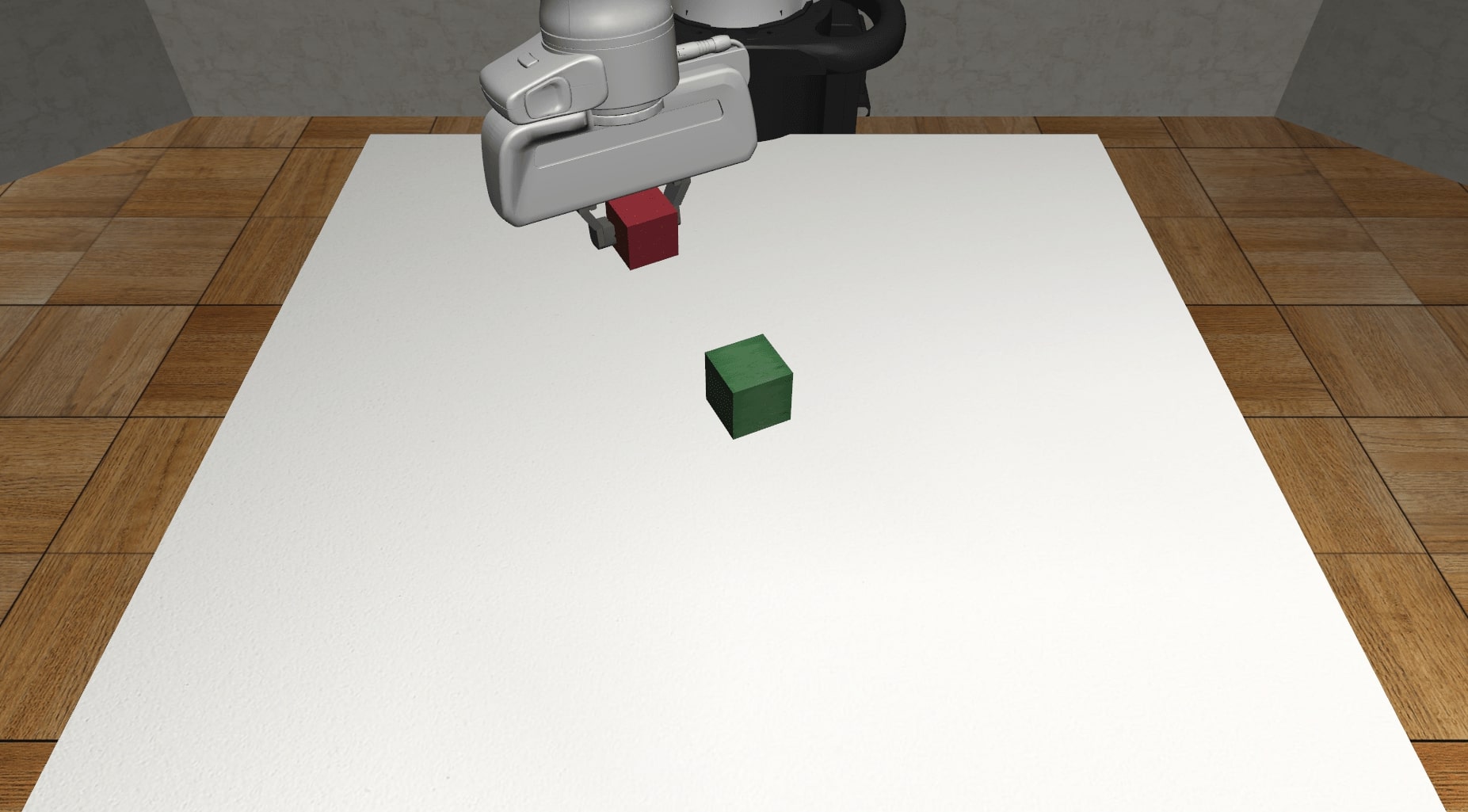}
        \vspace{-6mm}
        \caption*{\textbf{Stack}}
    \end{subfigure}
    \hfill
    \begin{subfigure}[t]{0.24\linewidth}
        \includegraphics[width=\textwidth]{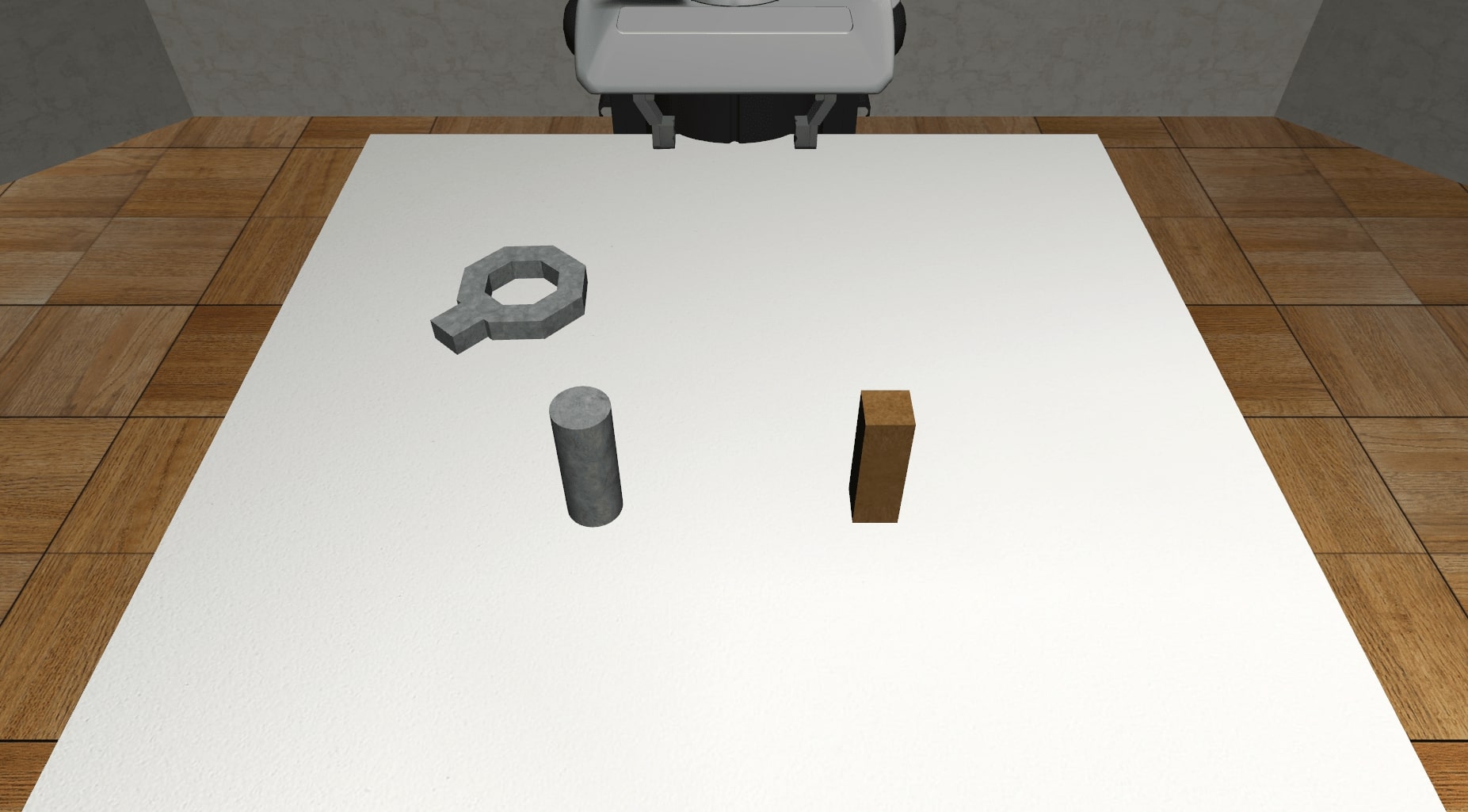}
        \vspace{-6mm}
        \caption*{\textbf{Nut Assembly}}
    \end{subfigure}
    \hfill
    \begin{subfigure}[t]{0.24\linewidth}
        \includegraphics[width=\textwidth]{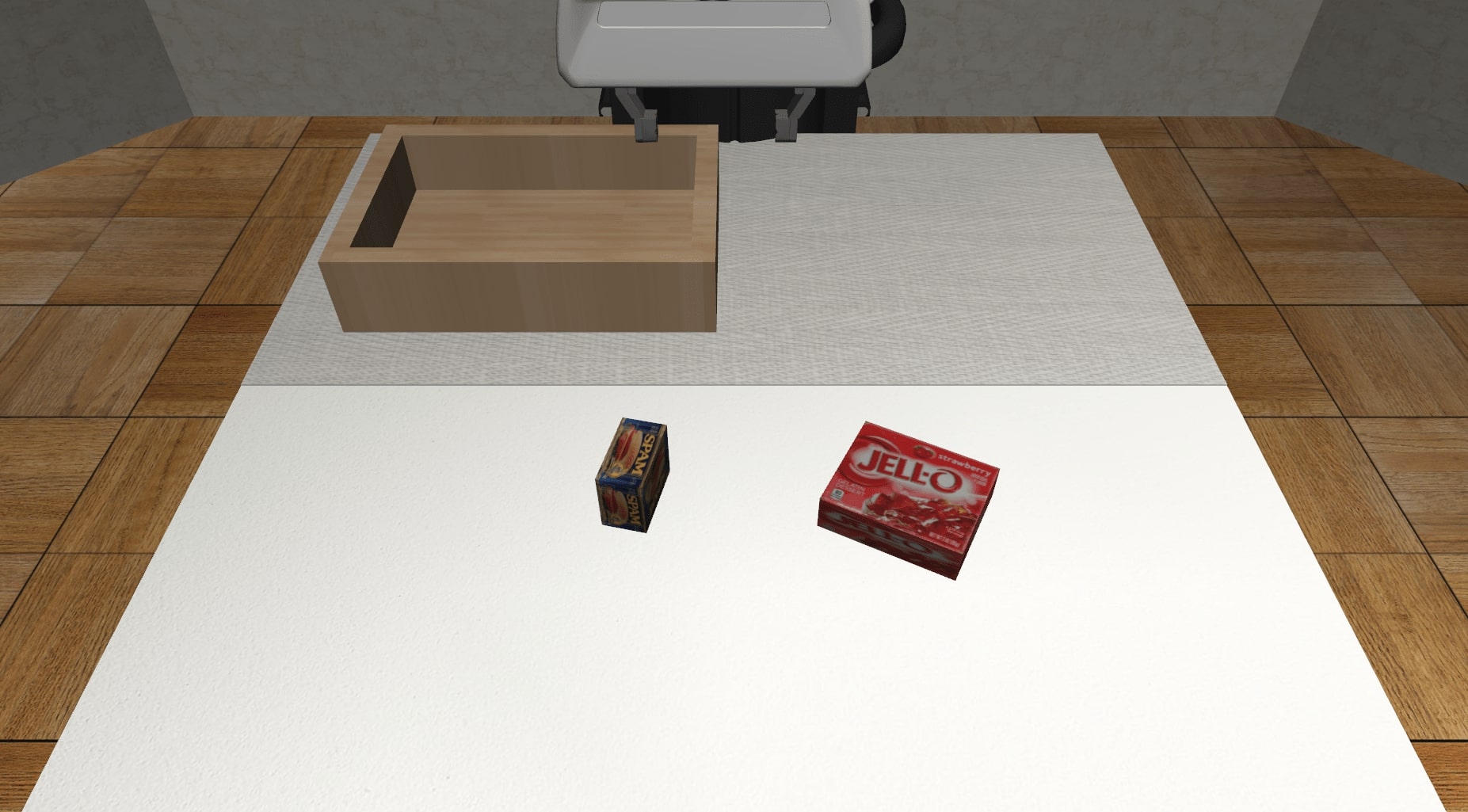}
        \vspace{-6mm}
        \caption*{\textbf{Cleanup}}
    \end{subfigure}
    \hfill
    \begin{subfigure}[t]{0.24\linewidth}
        \includegraphics[width=\textwidth]{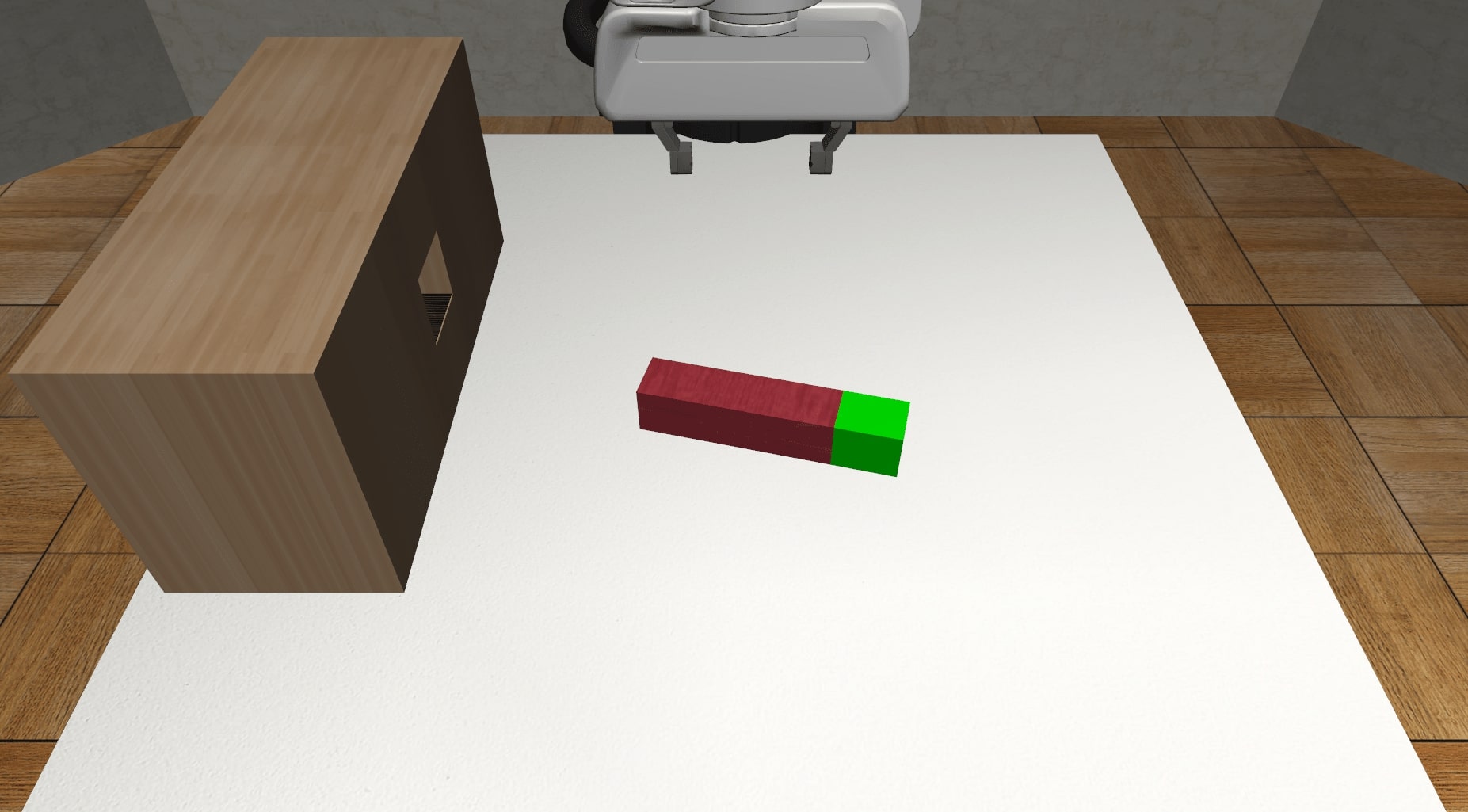}
        \vspace{-6mm}
        \caption*{\textbf{Peg Insertion}}
    \end{subfigure}
    \caption{\footnotesize{
    \textbf{Simulated Environments.} We perform evaluations on eight manipulation tasks. The first six come from the robosuite benchmark~\cite{robosuite2020}. We designed the last two to test our method in multi-stage, contact-rich tasks: Cleanup requires storing a spam can into a storage bin and a jello box at a corner; Peg Insertion requires inserting a peg into a block.
    }}
    \vspace{-16pt}
    \label{fig:envs}
\end{figure}

\begin{figure*}
    \vspace{2mm}
    \begin{subfigure}[b]{\textwidth}
    \centering
    \includegraphics[width=0.24\textwidth,valign=t]{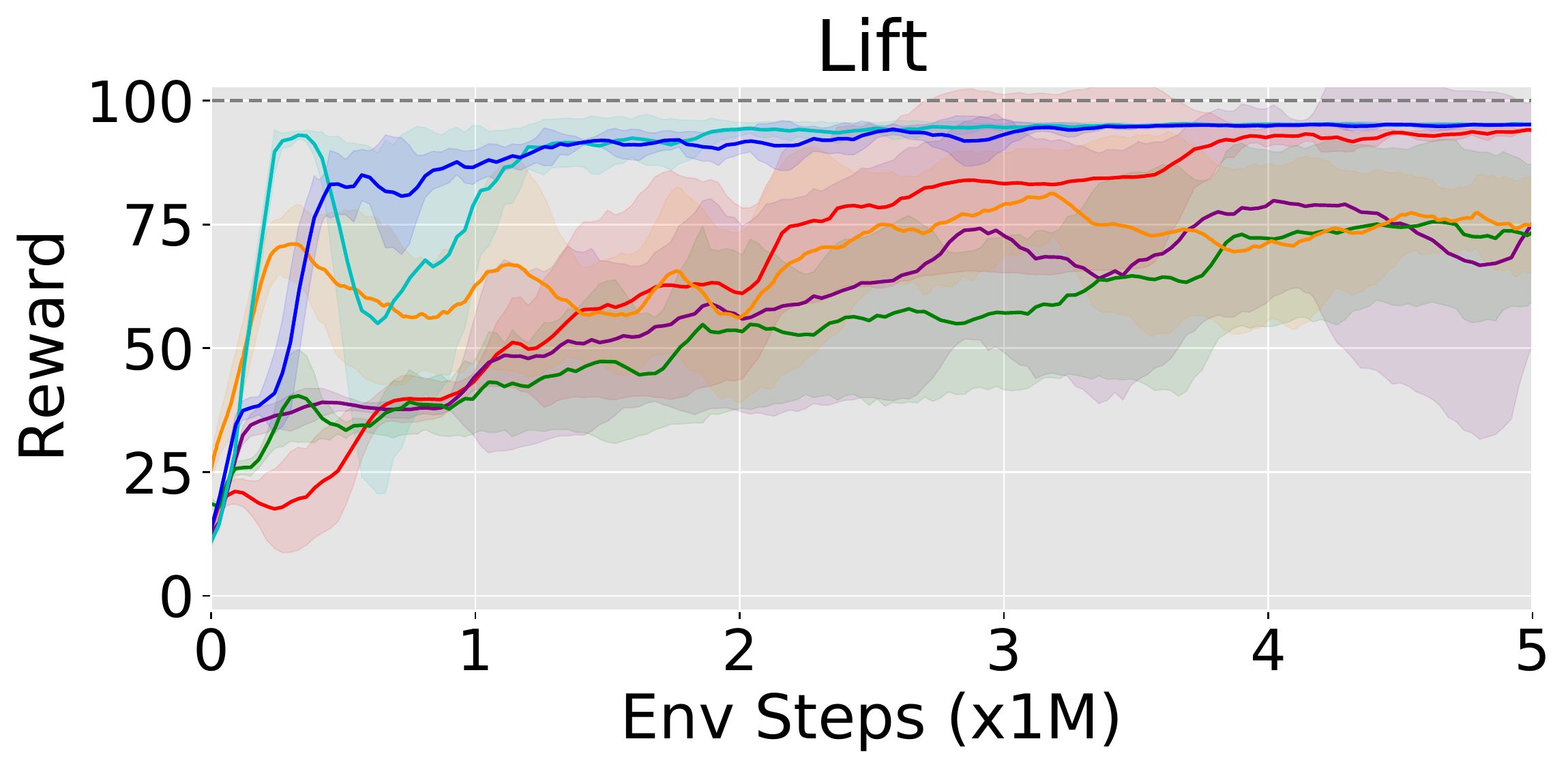}
    \hfill
    \includegraphics[width=0.24\textwidth,valign=t]{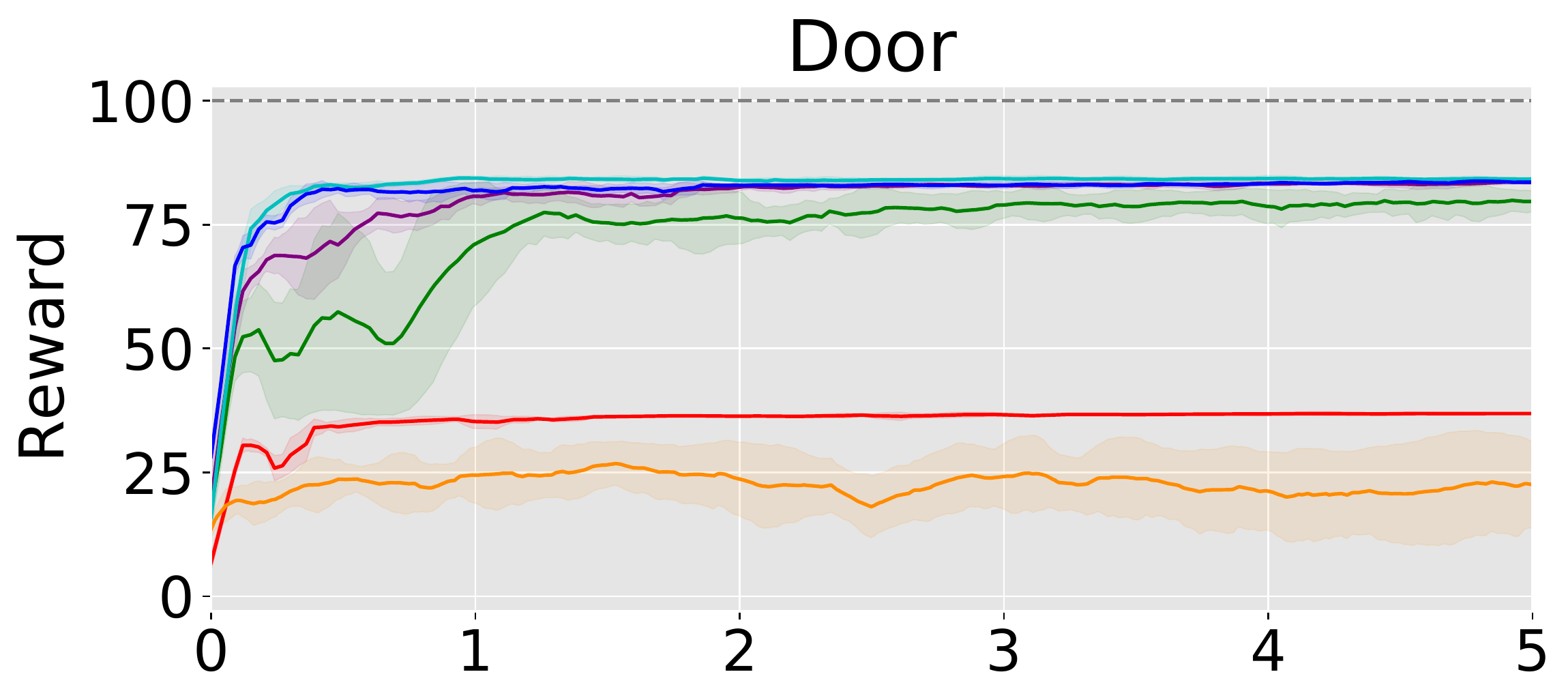}
    \hfill
    \includegraphics[width=0.24\textwidth,valign=t]{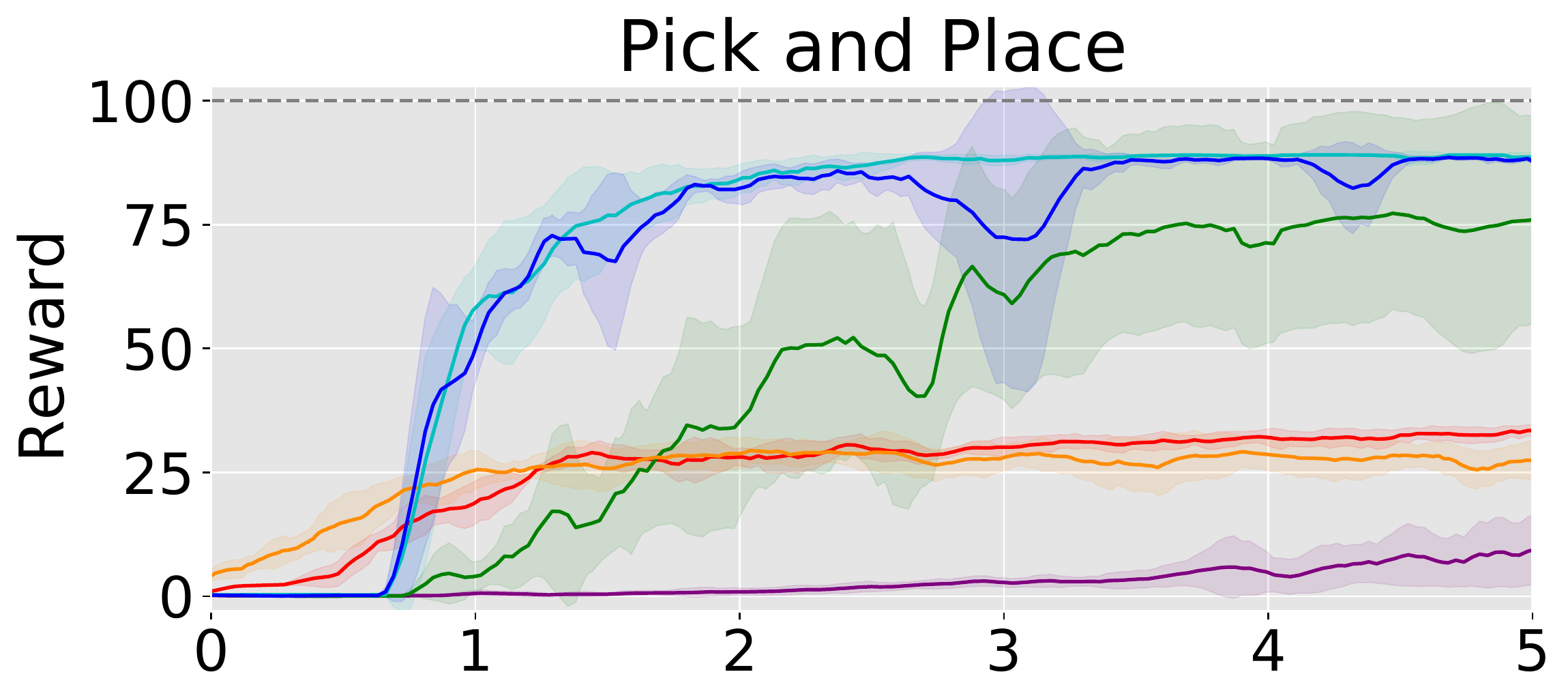}
    \hfill
    \includegraphics[width=0.24\textwidth,valign=t]{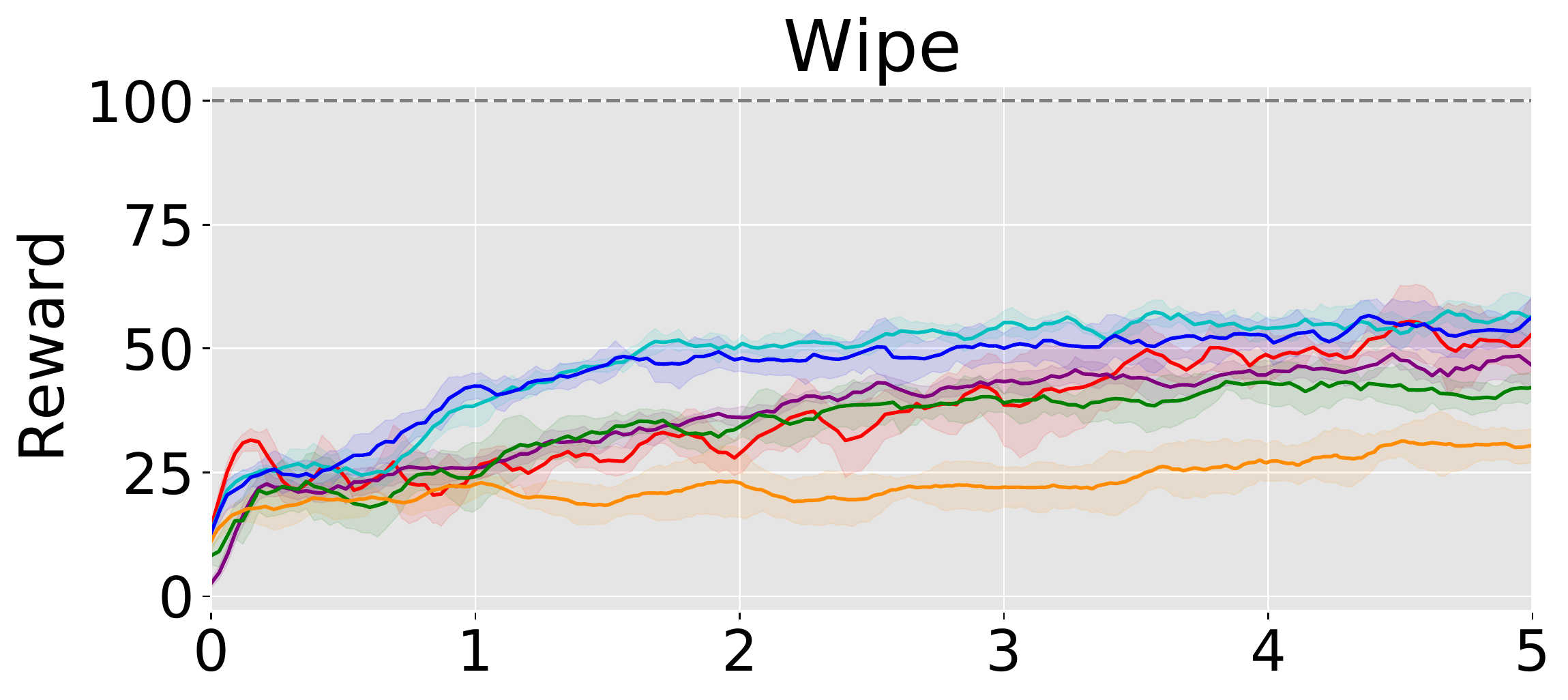}
    \includegraphics[width=0.24\textwidth]{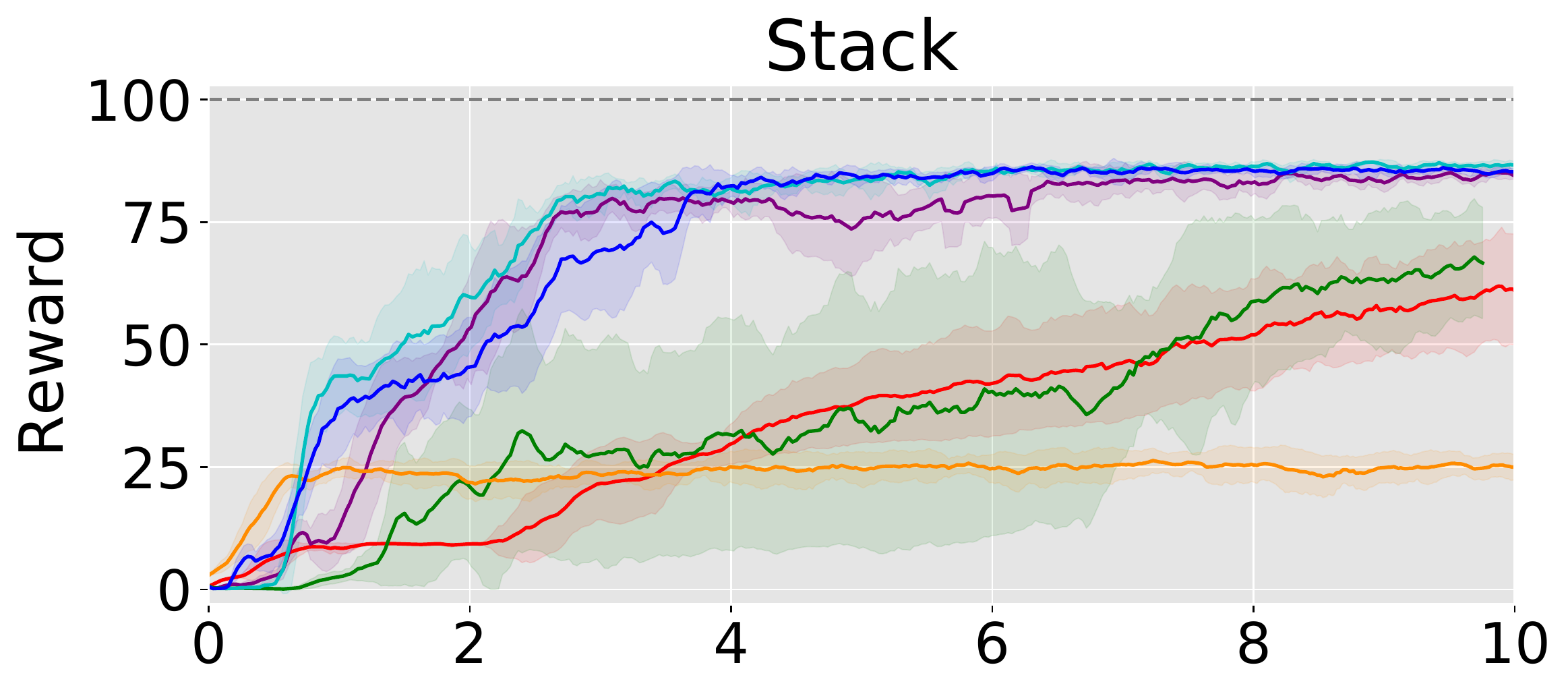}
    \hfill
    \includegraphics[width=0.24\textwidth]{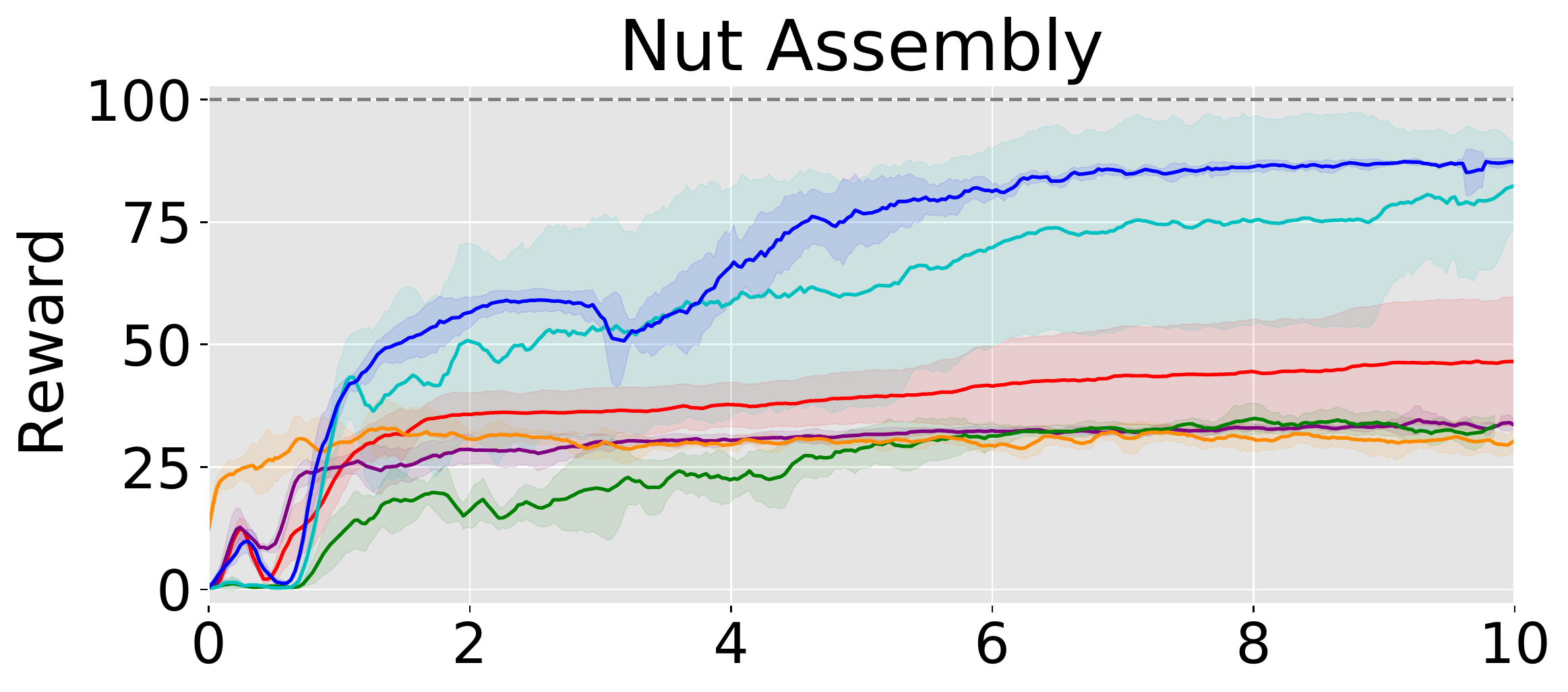}
    \hfill
    \includegraphics[width=0.24\textwidth]{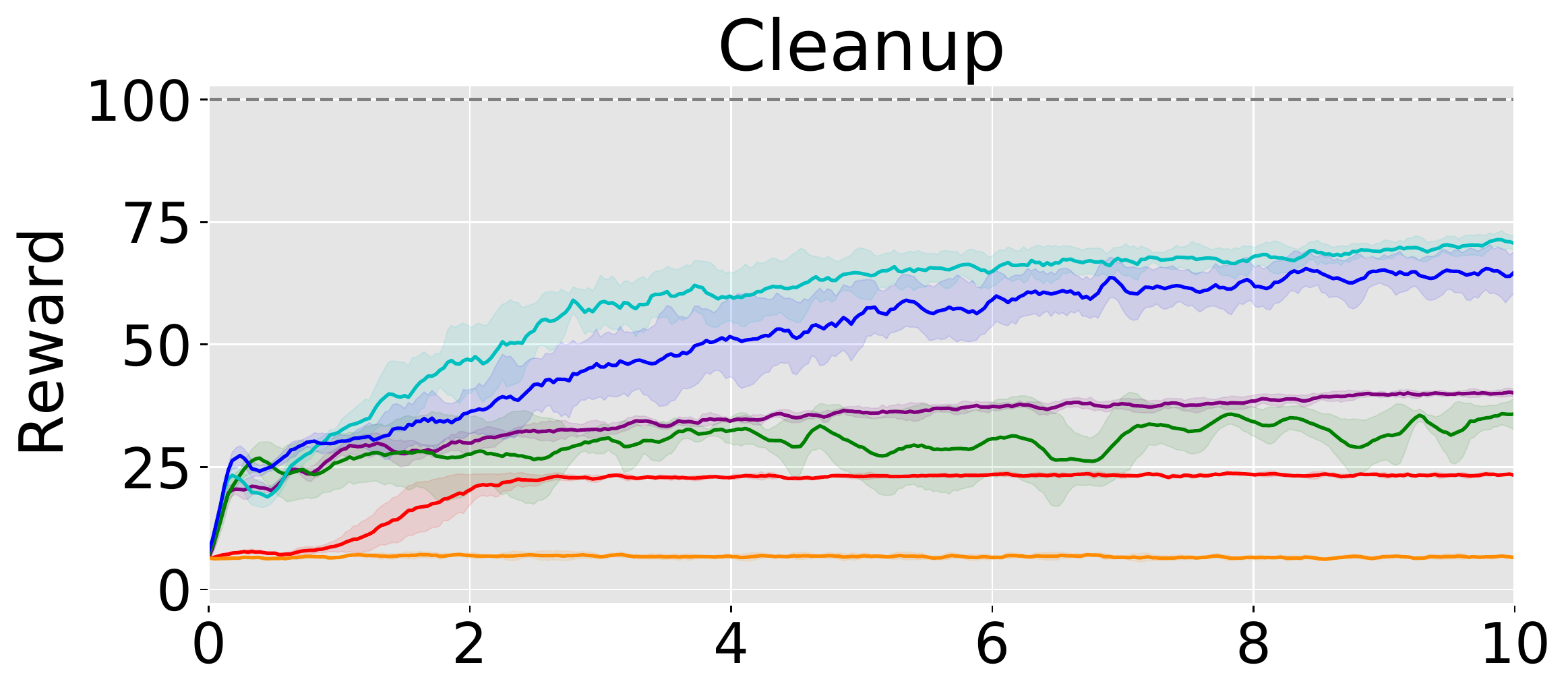}
    \hfill
    \vspace{0.10cm}
    \includegraphics[width=0.24\textwidth]{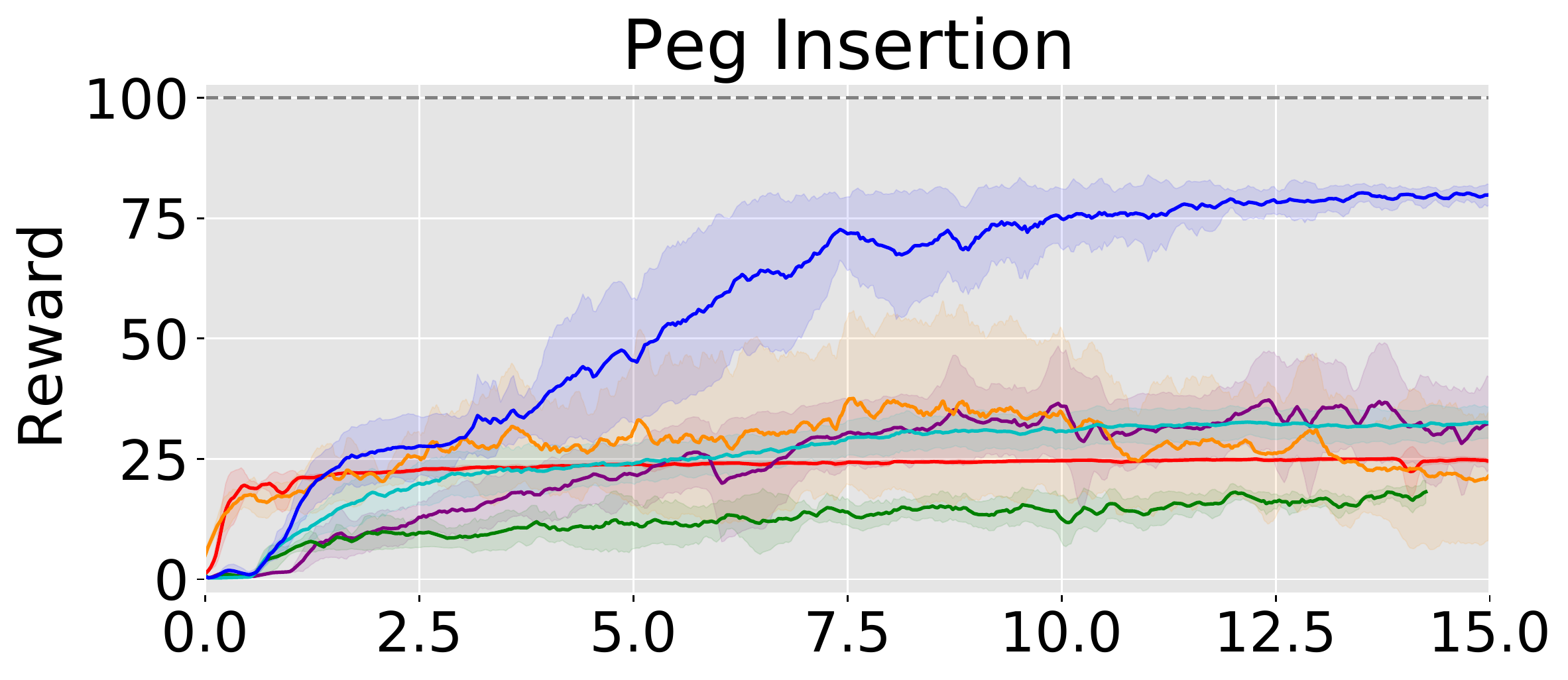}
    \vspace{0.25cm}
    \includegraphics[width=0.70\textwidth]{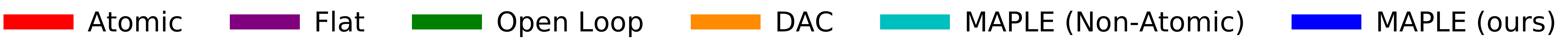}
    \vspace{-10pt}
    \centering
    \end{subfigure}
    \caption{
    \textbf{Main Results.} Learning curves showing average episodic task rewards throughout training, normalized between 0 and 100.
    All experiments are averaged over 5 seeds, with shaded regions depicting the standard deviation.
    }
    \label{plot:main-results}
\end{figure*}

\subsection{Quantifying Compositionality}
Our framework is based upon the hypothesis that most manipulation tasks have an intrinsic compositional structure and that our algorithm can discover this structure.
While there is no explicit mechanism to discover recurring patterns of primitives, our algorithm
exhibits compositional reasoning by preferring the use of high-level primitives over low-level ones.
Due to the temporal abstraction encapsulated by the high-level primitives, the agent can make far greater progress toward solving the task by using high-level primitives and thus receives higher average reward per timestep.
This incentivizes the agent to choose higher-level primitives over lower-level actions whenever appropriate.
We additionally examine the degree to which our learned agent exhibits compositional behaviors with a quantifiable metric.
Assuming a set of trajectories in which the agent solved a task $\cal{T}$: 
$\{\tau^i \}_{i=1}^n = \{ (s^i_1, (a^i_1, x^i_1), \cdots, s^i_{T_i}, (a^i_{T_i}, x^i_{T_i}), s^i_{T_i+1})\}_{i=1}^n$,
the \textit{task sketches} $\{K^i\}_{i=1}^n = \{(a^i_1, a^i_2, \cdots a^i_{T_i})\}_{i=1}^n$ capture high-level task semantics and provide useful abstractions through which we can analyze the compositional structure of these trajectories.
Intuitively, agents that demonstrate compositional reasoning will express recurring patterns of behaviors across their task sketches and prefer the use of high-level primitives over low-level ones.
We quantify this intuition by computing the \textit{Levenshtein distance}~\cite{levenshtein1966binary} among task sketches, which measures the minimum number of single-token edits needed to transform one task sketch to another.
We represent each non-atomic primitive \textit{type} as a unique token, and in order to explicitly discourage the use of low-level atomic actions, we represent each \textit{individual occurrence} of an atomic primitive in our task sketches as a unique token.
Given a task $\cal{T}$ and available primitives $\mathcal{L}$, we compute the compositionality of the agent's behavior as the average pairwise normalized score between the task sketches:
\begin{align}
f_{comp}(\cal{T}; \mathcal{L}) &= \frac{1}{n(n-1)} \sum_{i \neq j} 1 - \frac{d_{\text{Lev}}(K_i, K_j)}{\max(|K_i|, |K_j|)} \label{eq:comp-metric}
\end{align}
\section{Experiments}
\label{sec:experiments}
In our experiments we study 1) whether our method can compose pre-built behavior primitives and atomic actions to solve complex tasks,
2) the degree to which the learned behavior is compositional, and 3) whether our approach is amenable to transfer to task variants and to real hardware.

\subsection{Experimental Setup}
We examine these questions on robosuite~\cite{robosuite2020}, a framework for simulated robot manipulation tasks. We consider a comprehensive suite of eight manipulation tasks of varying complexities (see \cref{fig:envs}).
We adopt a Franka Emika Panda robot arm controlled via operational space control (OSC)~\cite{khatib1995osc}.
At each decision-making step the agent executes either an atomic OSC action or one of the non-atomic primitives outlined in \cref{sec:primitives}.
In return the agent receives 1) a dense reward signal indicating task progress and 2) an observation comprising the robot's proprioceptive state and pose information of the objects in the environment.
\subsection{Quantitative Evaluation}
We compare our method (\textbf{\METHODNOSPC}) to five baselines.
The first baseline uses exclusively atomic actions (\textbf{Atomic}), which corresponds to the standard Soft Actor-Critic model~\cite{haarnoja2018sac} trained on end-effector commands. To understand the effect of hierarchy on our policy design, we compare to a flat variant where the policy outputs the primitive type and parameters independently (\textbf{Flat}), following the design by~Lee et al.~\citet{lee2020skills} and~Neunert et al.~\citet{neunert2020continuousdiscrete}.
We also compare to a variant of our method using an open loop task policy (\textbf{Open Loop}), following Chitnis et al.~\citet{chitnis2020schema} which suggests utilizing an open-loop task schema improves the sample efficiency of the algorithm.
Next, we compare to Hierarchical Reinforcement learning with Off-policy correction (\textbf{HIRO})~\cite{nachum2018hiro} and Double Actor-Critic (\textbf{DAC})~\cite{zhang2019dac}, state-of-the-art hierarchical DRL methods which learn low-level policies (or options) along with high-level controllers.
HIRO failed to make progress and we thus omit it from our results. Finally, we compare to a self baseline where we include all primitives \textit{except} the atomic primitive (\textbf{\METHOD (Non-Atomic)}), to understand whether we need atomic actions to satisfy behaviors that cannot be fulfilled by the non-atomic primitives.
All baselines using behavior primitives use the affordance score outlined in \cref{sec:affordance}.

\begin{figure*}
    \vspace{1mm}
    \centering
    \includegraphics[width=1.0\linewidth]{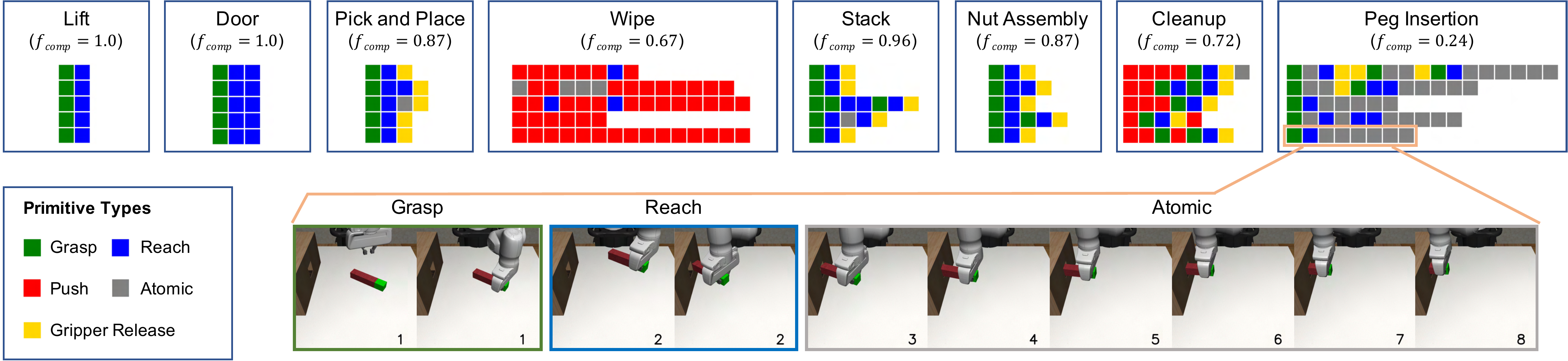}
    \caption{
    \textbf{Analyzing Learned Behavior.} (Top) We visualize the task sketches that our agent has learned.
    Each row corresponds to a single sketch progressing temporally from left to right.
    For each task we also report the compositionality score $f_{comp}$.
    (Bottom) We visualize the behavior for a peg insertion sketch.
    }
    \label{fig:task-sketches}
    \vspace{15pt}
\end{figure*}

\cref{plot:main-results} outlines environment rewards throughout training.
We also evaluated the final task success rates at the end of training: \METHOD achieved the highest average success rate across all baselines (90\%), compared to $19\%$ for the Atomic baseline, $36\%$ for Flat, $41\%$ for Open Loop, $11\%$ for DAC, and $79\%$ for \METHOD (Non-Atomic).
First, we see that the inclusion of non-atomic primitives allows \METHOD to significantly outperform the Atomic baseline, achieving on average 2-3$\times$ higher rewards and $71\%$ higher success rate.
Qualitatively we found that the Atomic baseline fails to advance past the first stage in most tasks while our method successfully solves all tasks.
Next we find that the Flat baseline is unable to reliably solve all tasks, demonstrating that our hierarchical policy design is key to reasoning over a heterogeneous set of primitives.
While the Open Loop baseline is able to solve basic tasks such as Door Opening and Pick and Place, it struggles with tasks that require the agent to adaptively reason about the current state of the task.
DAC is only able to solve the Lift task, highlighting the difficulty of learning complex tasks \textit{from scratch} even when employing temporal abstraction.
Finally we find that the Non-Atomic self baseline is on par with our method in most tasks, yet it notably fails for Peg Insertion as the non-atomic primitives are not expressive enough to perform the contact-rich insertion phase.
Together, these results highlight that given an appropriate primitive library and policy structure we can solve a wide range of manipulation tasks.

\subsection{Model Analysis}
\label{sec:comp-analysis}

\begin{figure}
    \setlength{\belowcaptionskip}{0pt}
    \centering
    \begin{subfigure}[t]{0.48\linewidth}
        \centering
        \includegraphics[width=1.0\linewidth]{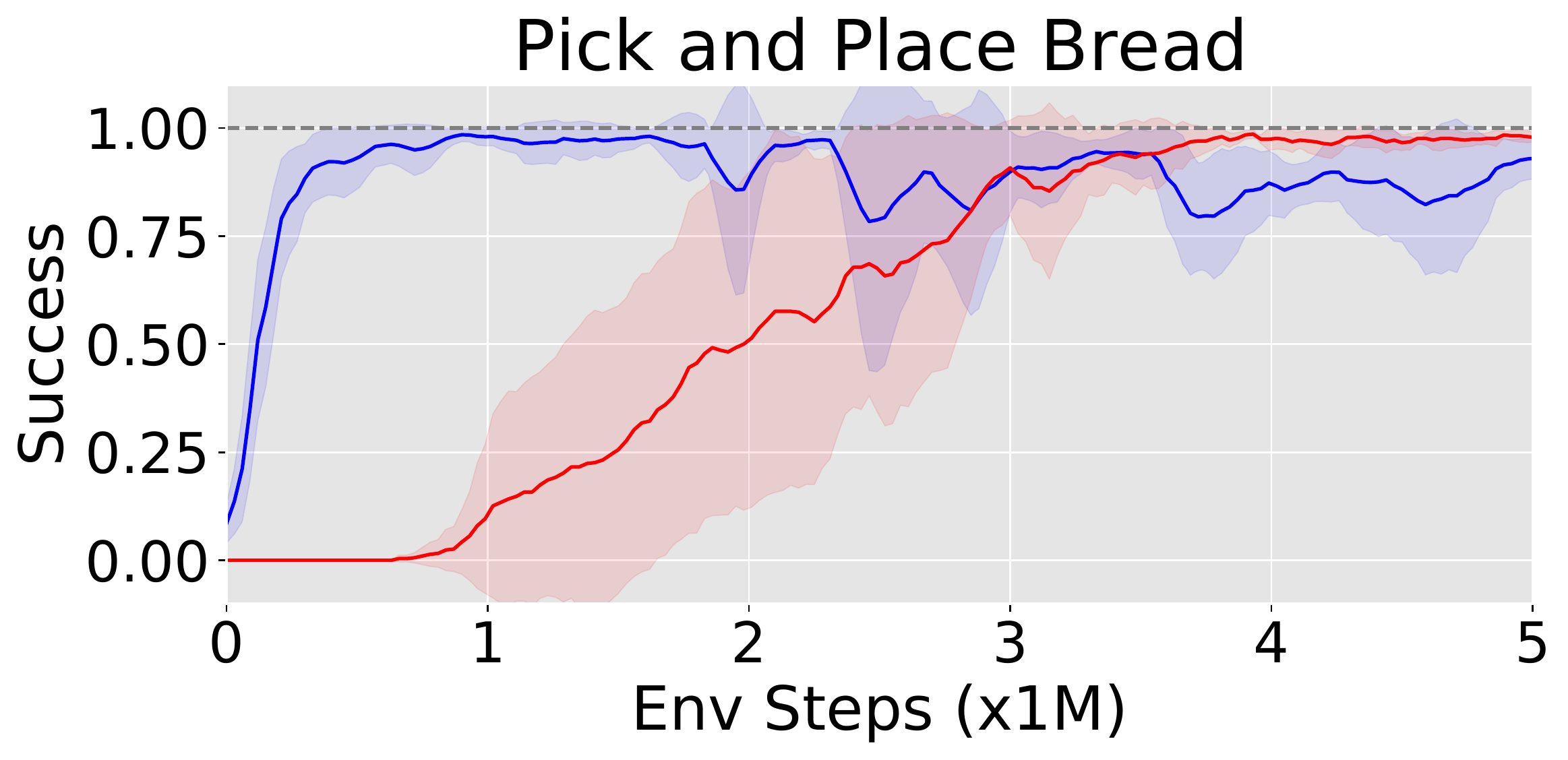}
        \includegraphics[width=0.95\linewidth]{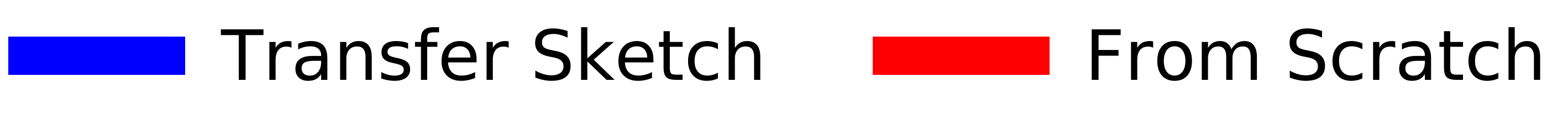}
        \vspace{-1pt}
        \caption{Task Transfer}
        \label{fig:transfer}
    \end{subfigure}
    \hfill
    \begin{subfigure}[t]{0.48\linewidth}
         \centering
         \includegraphics[width=1.0\linewidth]{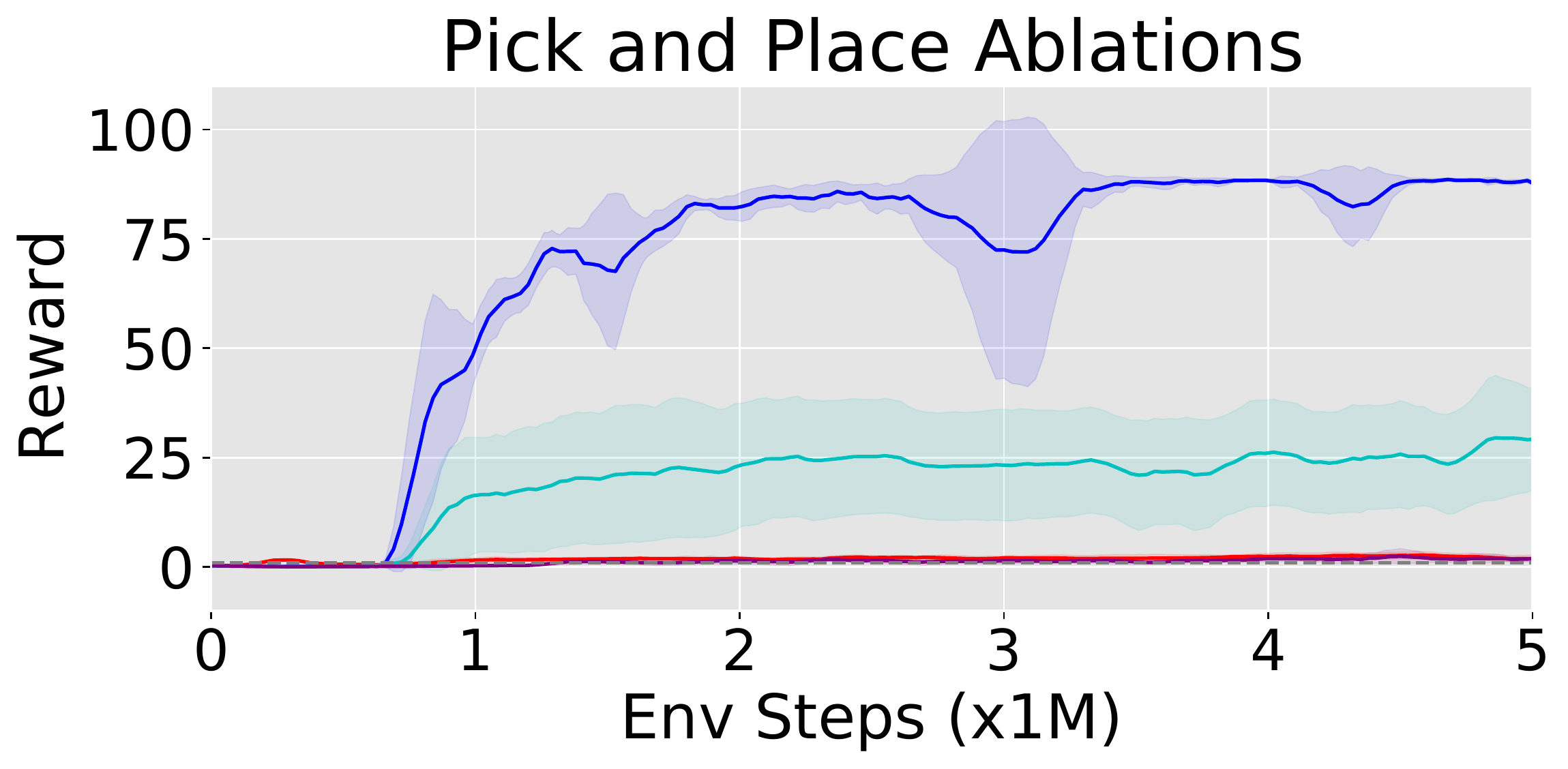}
        \includegraphics[width=0.60\linewidth]{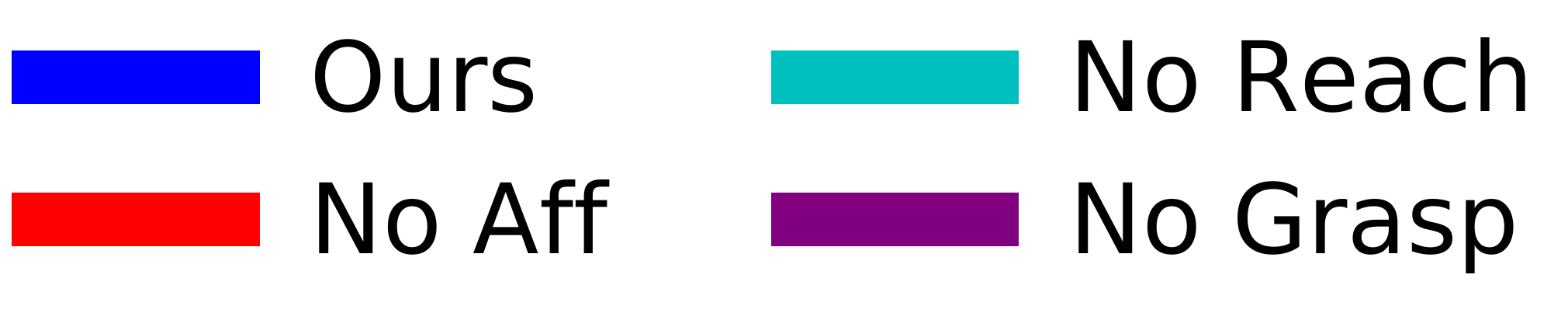}
        \vspace{-2pt}
        \caption{Ablations}
        \label{fig:ablations}
    \end{subfigure}
    \vspace{-1mm}
    \caption{
    \textbf{(a) Task Transfer.} We transfer the learned task sketch from a source task (pick and place can) to a semantically similar task variant (pick and place bread), enabling us to learn the target task over $5\times$ faster.
    \textbf{(b) Ablations}.
    Without affordances, reaching, or grasping, the agent is unable to solve tasks due to the exploration burden.
    }
    \vspace{-12pt}
\end{figure}

\noindent \textbf{Emergence of Compositional Structures.}
We present an analysis of the task sketches that our method learned for each task in \cref{fig:task-sketches}.
We see evidence that the agent unveils compositional task structures by applying temporally extended primitives whenever appropriate and relying on atomic actions otherwise.
For example, for the peg insertion task the agent leverages the grasping primitive to pick up the peg and the reaching primitive to align the peg with the hole in the block, but then it uses atomic actions for the contact-rich insertion phase.
In \cref{fig:task-sketches} we also quantify the compositionality of these task sketches via $f_{comp}$ defined in \cref{eq:comp-metric}. As expected, tasks involving contact interactions such as Peg Insertion and Wiping have lower scores than prehensile tasks such as Pick and Place and Stacking.

\noindent \textbf{Transfer to Semantically Similar Task Variants.}
We have seen how task sketches enable interpretability by serving as blueprints of high-level semantic task structure.
We can leverage these task sketches to accelerate learning on similar task instances.
We propose to re-use the task sketch from a semantically similar task, and only learn the corresponding primitive parameters.
We validate this idea on the Pick and Place domain, where we transfer the task sketch from a source task of placing a soda can into one bin, to a target task of placing a loaf of bread into a different bin. As shown in \cref{fig:transfer}, we solve the bread task significantly faster than learning the task from scratch with a sample efficiency of over $5\times$. This suggests that our task sketch serves as a high-level scaffold of a manipulation task, which can be re-used by learning algorithms for faster adaptation to related tasks.

\noindent \textbf{Ablation Study.}
We perform an ablation study examining the role of affordances and individual manipulation primitives in facilitating exploration.
We specifically perform experiments on the Pick and Place task, comparing our method (\textbf{Ours}) to ablations 1) without affordances in the reward function (\textbf{No Aff}), 2) without the reaching skill (\textbf{No Reach}), and 3) without the grasping skill (\textbf{No Grasp}).
We see in \cref{fig:ablations} that without these components the agent fails to solve the task, underscoring that our method is reliant on the appropriate primitive skills and affordances to effectively overcome the exploration burden.

\subsection{Real-World Evaluation}
\label{sec:real-robot}
\begin{figure}
    \centering
    \includegraphics[width=1.0\linewidth]{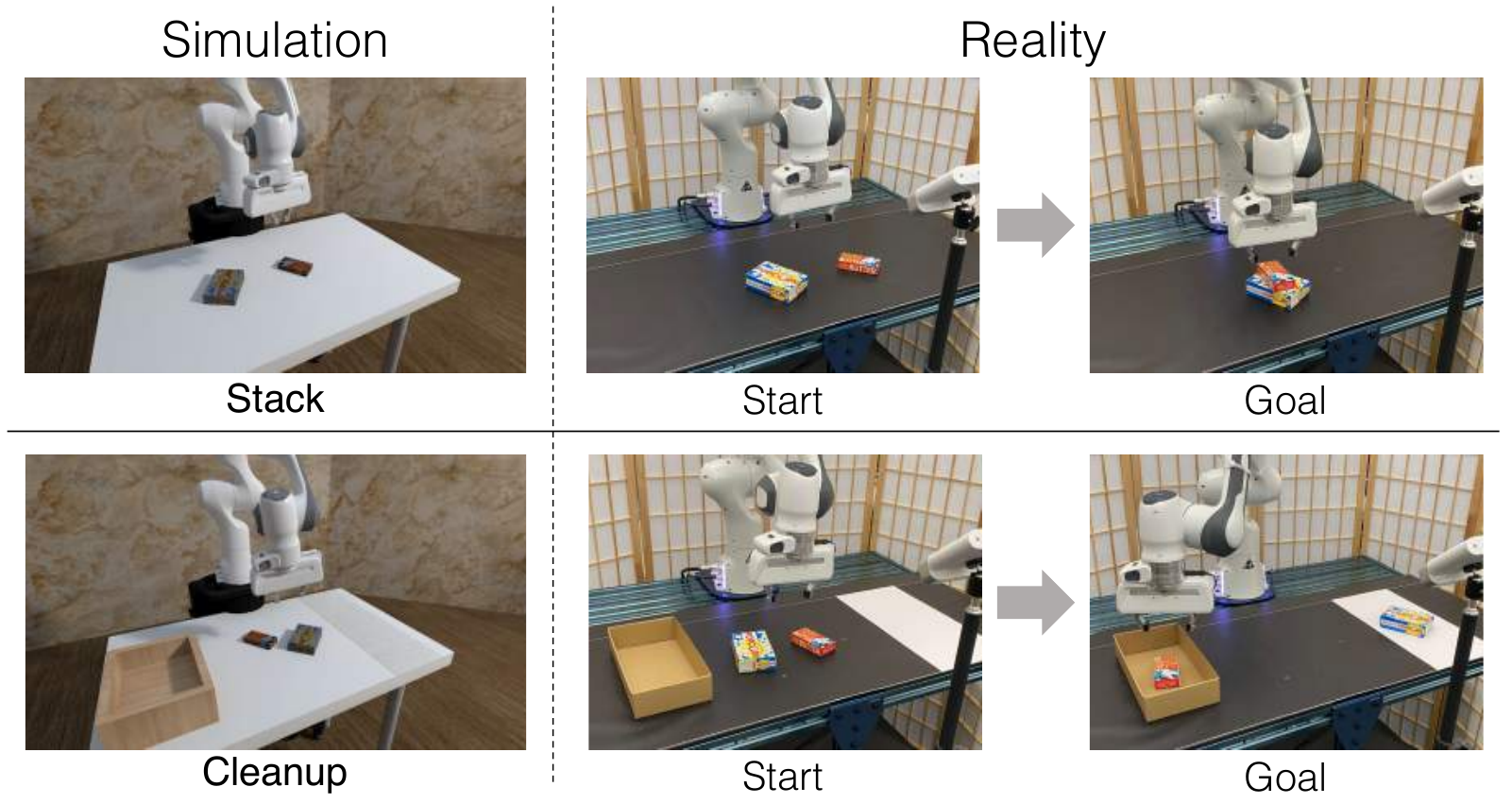}
    \caption{\textbf{Transfer to Real-World Tasks.} We transfer our policy trained on simulated environments to the real-world Stack and Cleanup tasks.}
    \label{fig:real-robot}
\end{figure}

We conclude with an evaluation on real-world copies of the Stack and Cleanup tasks (see \cref{fig:real-robot}). As our behavior primitives offer high-level action abstractions and encapsulate low-level complexities of motor actuation, our policies can directly transfer to the real world. We trained \METHOD on simulated versions of these tasks and executed the resulting policies to the real world.
We re-implemented our behavior primitives on the real robot and used an off-the-shelf pose estimation model~\cite{tremblay2018dope} to estimate environment states. We successfully transferred the policy to the real world, with an average success rate of $93\%$ on Stack and $83\%$ on Cleanup.

\section{Conclusion}
\label{sec:discussion}
We presented \METHODNOSPC, a reinforcement learning framework that incorporates behavior primitives in conjunction with low-level motor actions to solve complex manipulation tasks.
Our experiments demonstrate that behavior primitives can significantly improve exploration while low-level motor actions allow us to retain flexibility to learn intricate behaviors.
Our work opens the possibility for several avenues for future work.
First, learning affordances using data-driven methods~\cite{simeonov2020skills,kokic2020learning,mandikal2020graff,mo2021where2act} can expand the scalability of our method.
Second, while atomic actions can help fill in gaps where the primitives are insufficient (such as peg insertion), we are unable to fill in large gaps that require a significant number of low-level action executions (as seen in the ablation experiments).
Further research on exploration and credit assignment is needed to overcome these challenges.
Finally, an exciting avenue for future work is to continually discover recurring compositions of primitives and add them to the library of primitives, which can ultimately enable curriculum learning of progressively more challenging tasks.
\section*{ACKNOWLEDGMENT}
We would like to thank Yifeng Zhu for assisting with the real-world experiments and Abhishek Joshi for assisting with rendering content. We also thank Yifeng Zhu, Braham Snyder, and Josiah Wong for providing feedback on this manuscript.
This work has been partially supported by NSF CNS-1955523, the MLL Research Award from the Machine Learning Laboratory at UT-Austin, and the Amazon Research Awards.

\renewcommand*{\bibfont}{\footnotesize}
\printbibliography 

@inproceedings{haarnoja2018sac,
    title={Soft Actor-Critic: Off-Policy Maximum Entropy Deep Reinforcement Learning with a Stochastic Actor}, 
    author={Tuomas Haarnoja and Aurick Zhou and Pieter Abbeel and Sergey Levine},
    year={2018},
    booktitle={ICML},
}

@inproceedings{robosuite2020,
    title={robosuite: A Modular Simulation Framework and Benchmark for Robot Learning},
    author={Yuke Zhu and Josiah Wong and Ajay Mandlekar and Roberto Mart\'{i}n-Mart\'{i}n},
    booktitle={arXiv preprint arXiv:2009.12293},
    year={2020}
}

@inproceedings{masson2015reinforcement,
      title={Reinforcement Learning with Parameterized Actions}, 
      author={Warwick Masson and Pravesh Ranchod and George Konidaris},
      year={2016},
      booktitle={AAAI},
}

@inproceedings{hausknecht2016deep,
      title={Deep Reinforcement Learning in Parameterized Action Space}, 
      author={Matthew Hausknecht and Peter Stone},
      year={2016},
      booktitle={ICLR},
}

@article{wei2018hierarchical,
  title={Hierarchical Approaches for Reinforcement Learning in Parameterized Action Space},
  author={E. Wei and Drew Wicke and S. Luke},
  journal={arXiv},
  year={2018},
  volume={abs/1810.09656}
}

@article{xiong2018parametrized,
  title={Parametrized Deep Q-Networks Learning: Reinforcement Learning with Discrete-Continuous Hybrid Action Space},
  author={J. Xiong and Qing Wang and Zhuoran Yang and Peng Sun and Lei Han and Yang Zheng and Haobo Fu and T. Zhang and Ji Liu and Han Liu},
  journal={arXiv},
  year={2018},
  volume={abs/1810.06394}
}

@inproceedings{fan2019hybrid,
      title={Hybrid Actor-Critic Reinforcement Learning in Parameterized Action Space}, 
      author={Zhou Fan and Rui Su and Weinan Zhang and Yong Yu},
      year={2019},
      booktitle={IJCAI},
}

@inproceedings{yamada2020mopa,
  title={Motion Planner Augmented Reinforcement Learning for Obstructed Environments},
  author={Jun Yamada and Youngwoon Lee and Gautam Salhotra and Karl Pertsch and Max Pflueger and Gaurav S. Sukhatme and Joseph J. Lim and Peter Englert},
  booktitle={CoRL},
  year={2020},
}

@inproceedings{xiali2020relmogen, 
    title={ReLMoGen: Leveraging Motion Generation in Reinforcement Learning for Mobile Manipulation},
    author={Xia, Fei and Li, Chengshu and Mart{\'\i}n-Mart{\'\i}n, Roberto and Litany, Or and Toshev, Alexander and Savarese, Silvio }, 
    booktitle={ICRA},
    year={2021} 
}

@inproceedings{pertsch2020spirl,
    title={Accelerating Reinforcement Learning with Learned Skill Priors},
    author={Karl Pertsch and Youngwoon Lee and Joseph J. Lim},
    booktitle={CoRL},
    year={2020},
}

@inproceedings{singh2020parrot,
      title={Parrot: Data-Driven Behavioral Priors for Reinforcement Learning}, 
      author={Avi Singh and Huihan Liu and Gaoyue Zhou and Albert Yu and Nicholas Rhinehart and Sergey Levine},
      year={2020},
      booktitle={ICLR}
}

@inproceedings{ajay2020opal,
      title={OPAL: Offline Primitive Discovery for Accelerating Offline Reinforcement Learning}, 
      author={Anurag Ajay and Aviral Kumar and Pulkit Agrawal and Sergey Levine and Ofir Nachum},
      year={2021},
      booktitle={ICLR}
}

@misc{schulman2017ppo,
      title={Proximal Policy Optimization Algorithms}, 
      author={John Schulman and Filip Wolski and Prafulla Dhariwal and Alec Radford and Oleg Klimov},
      year={2017},
      eprint={1707.06347},
      archivePrefix={arXiv},
      primaryClass={cs.LG}
}

@inproceedings{bellemare2016unifying,
      title={Unifying Count-Based Exploration and Intrinsic Motivation}, 
      author={Marc G. Bellemare and Sriram Srinivasan and Georg Ostrovski and Tom Schaul and David Saxton and Remi Munos},
      year={2016},
      booktitle={NIPS},
}

@article{garrett2020tamp,
      title={Integrated Task and Motion Planning},
      author={Caelan Reed Garrett and Rohan Chitnis and Rachel Holladay and Beomjoon Kim and Tom Silver and Leslie Pack Kaelbling and Tomás Lozano-Pérez},
      year={2021},
      journal = {Annual Review of Control, Robotics, and Autonomous Systems},
}

@inproceedings{maddison2017concrete,
      title={The Concrete Distribution: A Continuous Relaxation of Discrete Random Variables}, 
      author={Chris J. Maddison and Andriy Mnih and Yee Whye Teh},
      year={2017},
      booktitle={ICLR},
}

@inproceedings{jang2017categorical,
      title={Categorical Reparameterization with Gumbel-Softmax}, 
      author={Eric Jang and Shixiang Gu and Ben Poole},
      year={2017},
      booktitle={ICLR},
}

@inproceedings{kaelbling2011hpn,
  author={Leslie P. Kaelbling and Tomás Lozano-Pérez},
  title={Hierarchical task and motion planning in the now},
  booktitle={ICRA}, 
  year={2011},
}

@inproceedings{nachum2018hiro,
      title={Data-Efficient Hierarchical Reinforcement Learning}, 
      author={Ofir Nachum and Shixiang Gu and Honglak Lee and Sergey Levine},
      year={2018},
      booktitle={NeurIPS},
}

@article{karaman2011rrtstar,
  title={Sampling-based algorithms for optimal motion planning},
  author={Sertac Karaman and Emilio Frazzoli},
  journal={IJRR},
  year={2011},
}

@article{osa2018il,
  title={An Algorithmic Perspective on Imitation Learning},
  author={Takayuki Osa and J. Pajarinen and G. Neumann and J. Bagnell and P. Abbeel and Jan Peters},
  journal={Foundations and Trends in Robotics},
  year={2018},
}

@article{wang2021comptamp,
      title={Learning compositional models of robot skills for task and motion planning}, 
      author={Zi Wang and Caelan Reed Garrett and Leslie Pack Kaelbling and Tomás Lozano-Pérez},
      journal={IJRR},
      year={2021},
}

@inproceedings{fang2018demo2vec,
      title={Demo2Vec: Reasoning Object Affordances from Online Videos}, 
      author={Kuan Fang and Te-Lin Wu and Daniel Yang and Silvio Savarese and Joseph J. Lim},
      year={2018},
      booktitle={CVPR},
}

@inproceedings{mandikal2020graff,
  title = {Learning Dexterous Grasping with Object-Centric Visual Affordances},
  author = {Mandikal, Priyanka and Grauman, Kristen},
  booktitle = {ICRA},
  year = {2021}
}

@inproceedings{nagarajan2019grounded,
      title={Grounded Human-Object Interaction Hotspots from Video}, 
      author={Tushar Nagarajan and Christoph Feichtenhofer and Kristen Grauman},
      year={2019},
      booktitle={ICCV},
}

@inproceedings{nagarajan2020learning,
      title={Learning Affordance Landscapes for Interaction Exploration in 3D Environments},
      author={Tushar Nagarajan and Kristen Grauman},
      year={2020},
      booktitle={NeurIPS},
}

@ARTICLE{levenshtein1966binary,
       author = {{Levenshtein}, V.~I.},
        title = "{Binary Codes Capable of Correcting Deletions, Insertions and Reversals}",
      journal = {Soviet Physics Doklady},
         year = 1966,
        month = feb,
       volume = {10},
        pages = {707},
       adsurl = {https://ui.adsabs.harvard.edu/abs/1966SPhD...10..707L}
}

@inproceedings{allshire2021laser,
      title={LASER: Learning a Latent Action Space for Efficient Reinforcement Learning}, 
      author={Arthur Allshire and Roberto Martín-Martín and Charles Lin and Shawn Manuel and Silvio Savarese and Animesh Garg},
      year={2021},
      booktitle={ICRA},
}

@misc{openai2019solving,
      title={Solving Rubik's Cube with a Robot Hand}, 
      author={OpenAI and Ilge Akkaya and Marcin Andrychowicz and Maciek Chociej and Mateusz Litwin and Bob McGrew and Arthur Petron and Alex Paino and Matthias Plappert and Glenn Powell and Raphael Ribas and Jonas Schneider and Nikolas Tezak and Jerry Tworek and Peter Welinder and Lilian Weng and Qiming Yuan and Wojciech Zaremba and Lei Zhang},
      year={2019},
      eprint={1910.07113},
      archivePrefix={arXiv},
      primaryClass={cs.LG}
}

@inproceedings{nair2018overcoming,
      title={Overcoming Exploration in Reinforcement Learning with Demonstrations}, 
      author={Ashvin Nair and Bob McGrew and Marcin Andrychowicz and Wojciech Zaremba and Pieter Abbeel},
      year={2018},
      booktitle={ICRA},
}

@inproceedings{rajeswaran2018learning,
      title={Learning Complex Dexterous Manipulation with Deep Reinforcement Learning and Demonstrations}, 
      author={Aravind Rajeswaran and Vikash Kumar and Abhishek Gupta and Giulia Vezzani and John Schulman and Emanuel Todorov and Sergey Levine},
      year={2018},
      booktitle={RSS}
}

@inproceedings{mahler2017dexnet2,
      title={Dex-Net 2.0: Deep Learning to Plan Robust Grasps with Synthetic Point Clouds and Analytic Grasp Metrics}, 
      author={Jeffrey Mahler and Jacky Liang and Sherdil Niyaz and Michael Laskey and Richard Doan and Xinyu Liu and Juan Aparicio Ojea and Ken Goldberg},
      year={2017},
      booktitle={RSS},
}

@inproceedings{neunert2020continuousdiscrete,
      title={Continuous-Discrete Reinforcement Learning for Hybrid Control in Robotics}, 
      author={Michael Neunert and Abbas Abdolmaleki and Markus Wulfmeier and Thomas Lampe and Jost Tobias Springenberg and Roland Hafner and Francesco Romano and Jonas Buchli and Nicolas Heess and Martin Riedmiller},
      year={2019},
      booktitle={CoRL},
}

@inproceedings{xu2018neural,
      title={Neural Task Programming: Learning to Generalize Across Hierarchical Tasks}, 
      author={Danfei Xu and Suraj Nair and Yuke Zhu and Julian Gao and Animesh Garg and Li Fei-Fei and Silvio Savarese},
      year={2018},
      booktitle={ICRA},
}

@inproceedings{huang2019neural,
      title={Neural Task Graphs: Generalizing to Unseen Tasks from a Single Video Demonstration}, 
      author={De-An Huang and Suraj Nair and Danfei Xu and Yuke Zhu and Animesh Garg and Li Fei-Fei and Silvio Savarese and Juan Carlos Niebles},
      year={2019},
      booktitle={CVPR},
}

@inproceedings{gupta2019relay,
      title={Relay Policy Learning: Solving Long-Horizon Tasks via Imitation and Reinforcement Learning}, 
      author={Abhishek Gupta and Vikash Kumar and Corey Lynch and Sergey Levine and Karol Hausman},
      year={2019},
      booktitle={CoRL},
}

@inproceedings{sharma2020dynamicsaware,
      title={Dynamics-Aware Unsupervised Discovery of Skills}, 
      author={Archit Sharma and Shixiang Gu and Sergey Levine and Vikash Kumar and Karol Hausman},
      year={2020},
      booktitle={ICLR},
}

@InProceedings{coreyes18sectar,
  title = 	 {Self-Consistent Trajectory Autoencoder: Hierarchical Reinforcement Learning with Trajectory Embeddings},
  author =       {Co-Reyes, John and Liu, YuXuan and Gupta, Abhishek and Eysenbach, Benjamin and Abbeel, Pieter and Levine, Sergey},
  booktitle = 	 {ICML},
  year = 	 {2018},
}

@inproceedings{bacon2016optioncritic,
  title={The Option-Critic Architecture},
  author={Pierre-Luc Bacon and Jean Harb and Doina Precup},
  booktitle={AAAI},
  year={2017}
}

@inproceedings{pathak2017curiositydriven,
    Author = {Pathak, Deepak and
    Agrawal, Pulkit and
    Efros, Alexei A. and
    Darrell, Trevor},
    Title = {Curiosity-driven Exploration
    by Self-supervised Prediction},
    Booktitle = {ICML},
    Year = {2017}
}

@inproceedings{houthooft2017vime,
  title={VIME: Variational Information Maximizing Exploration},
  author={Rein Houthooft and Xi Chen and Yan Duan and J. Schulman and F. Turck and P. Abbeel},
  booktitle={NIPS},
  year={2016}
}

@inproceedings{pathak2019selfsupervised,
  Author = {Pathak, Deepak and
  Gandhi, Dhiraj and Gupta, Abhinav},
  Title = {Self-Supervised Exploration
  via Disagreement},
  Booktitle = {ICML},
  Year = {2019}
}

@inproceedings{eysenbach2018diversity,
      title={Diversity is All You Need: Learning Skills without a Reward Function}, 
      author={Benjamin Eysenbach and Abhishek Gupta and Julian Ibarz and Sergey Levine},
      year={2018},
      booktitle={ICLR}
}

@article{ijspeert2013dmps,
  author={Ijspeert, Auke Jan and Nakanishi, Jun and Hoffmann, Heiko and Pastor, Peter and Schaal, Stefan},
  journal={Neural Computation}, 
  title={Dynamical Movement Primitives: Learning Attractor Models for Motor Behaviors}, 
  year={2013},
 }

@inproceedings{fujimoto2019offpolicy,
      title={Off-Policy Deep Reinforcement Learning without Exploration}, 
      author={Scott Fujimoto and David Meger and Doina Precup},
      Booktitle = {ICML},
      Year = {2019}
}

@inproceedings{kumar2020conservative,
      title={Conservative Q-Learning for Offline Reinforcement Learning}, 
      author={Aviral Kumar and Aurick Zhou and George Tucker and Sergey Levine},
      year={2020},
      booktitle={NeurIPS},
}

@article{khatib1995osc,
    author = {Oussama Khatib},
    title ={Inertial Properties in Robotic Manipulation: An Object-Level Framework},
    journal = {IJRR},
    year = {1995},
}

@misc{fu2021d4rl,
      title={D4RL: Datasets for Deep Data-Driven Reinforcement Learning}, 
      author={Justin Fu and Aviral Kumar and Ofir Nachum and George Tucker and Sergey Levine},
      year={2020},
}

@article{bohg2014graspsurvey,
  title={Data-driven grasp synthesis—a survey},
  author={Bohg, Jeannette and Morales, Antonio and Asfour, Tamim and Kragic, Danica},
  journal={IEEE Transactions on Robotics},
  year={2013},
  publisher={IEEE}
}

@inproceedings{lee2020guapo,
  title={Guided Uncertainty-Aware Policy Optimization: Combining Learning and Model-Based Strategies for Sample-Efficient Policy Learning},
  author={Michelle A. Lee and Carlos Florensa and Jonathan Tremblay and Nathan Ratliff and Animesh Garg and Fabio Ramos and Dieter Fox},
  booktitle={ICRA},
  year={2020},
}

@inproceedings{mandlekar2020iris,
      title={IRIS: Implicit Reinforcement without Interaction at Scale for Learning Control from Offline Robot Manipulation Data}, 
      author={Ajay Mandlekar and Fabio Ramos and Byron Boots and Silvio Savarese and Li Fei-Fei and Animesh Garg and Dieter Fox},
      year={2020},
      booktitle={ICRA},
}

@inproceedings{eysenbach2019sorb,
      title={Search on the Replay Buffer: Bridging Planning and Reinforcement Learning}, 
      author={Benjamin Eysenbach and Ruslan Salakhutdinov and Sergey Levine},
      year={2019},
      booktitle={NeurIPS},
}

@inproceedings{nasiriany2019planning,
      title={Planning with Goal-Conditioned Policies}, 
      author={Soroush Nasiriany and Vitchyr H. Pong and Steven Lin and Sergey Levine},
      year={2019},
      booktitle={NeurIPS}
}

@inproceedings{amato1996randomized,
  author={Amato, N.M. and Wu, Y.},
  booktitle={ICRA}, 
  title={A randomized roadmap method for path and manipulation planning}, 
  year={1996},
}

@inproceedings{kalashnikov2018qtopt,
      title={QT-Opt: Scalable Deep Reinforcement Learning for Vision-Based Robotic Manipulation}, 
      author={Dmitry Kalashnikov and Alex Irpan and Peter Pastor and Julian Ibarz and Alexander Herzog and Eric Jang and Deirdre Quillen and Ethan Holly and Mrinal Kalakrishnan and Vincent Vanhoucke and Sergey Levine},
      year={2018},
      booktitle={CoRL},
}

@article{kalashnikov2021mtopt,
      title={MT-Opt: Continuous Multi-Task Robotic Reinforcement Learning at Scale}, 
      author={Dmitry Kalashnikov and Jacob Varley and Yevgen Chebotar and Benjamin Swanson and Rico Jonschkowski and Chelsea Finn and Sergey Levine and Karol Hausman},
      year={2021},
      journal={arXiv},
}

@misc{openai2019learning,
      title={Learning Dexterous In-Hand Manipulation}, 
      author={OpenAI and Marcin Andrychowicz and Bowen Baker and Maciek Chociej and Rafal Jozefowicz and Bob McGrew and Jakub Pachocki and Arthur Petron and Matthias Plappert and Glenn Powell and Alex Ray and Jonas Schneider and Szymon Sidor and Josh Tobin and Peter Welinder and Lilian Weng and Wojciech Zaremba},
      year={2019},
      eprint={1808.00177},
      archivePrefix={arXiv},
      primaryClass={cs.LG}
}

@inproceedings{zhang2019dac,
      title={DAC: The Double Actor-Critic Architecture for Learning Options}, 
      author={Shangtong Zhang and Shimon Whiteson},
      year={2019},
      booktitle={NeurIPS}
}

@inproceedings{bagaria2020dsc,
      title={Option Discovery using Deep Skill Chaining}, 
      author={Akhil Bagaria and George Konidaris},
      year={2020},
      booktitle={ICLR}
}

@inproceedings{smith2018inference,
  title={An inference-based policy gradient method for learning options},
  author={Smith, Matthew and Hoof, Herke and Pineau, Joelle},
  booktitle={ICML},
  year={2018},
}

@inproceedings{chitnis2020schema,
  title={Efficient Bimanual Manipulation Using Learned Task Schemas},
  author={Rohan Chitnis and Shubham Tulsiani and Saurabh Gupta and Abhinav Gupta},
  booktitle={ICRA},
  year={2020},
}

@inproceedings{strudel2020skills,
  title={Learning to combine primitive skills: A step towards versatile robotic manipulation},
  author={Robin Strudel and Alexander Pashevich and Igor Kalevatykh and Ivan Laptev and Josef Sivic and Cordelia Schmid},
  booktitle={ICRA},
  year={2020},
}

@inproceedings{sharma2020skills,
  title={Learning to compose hierarchical object-centric controllers for robotic manipulation},
  author={Mohit Sharma and Jacky Liang and Jialiang Zhao and Alex LaGrassa and Oliver Kroemer},
  booktitle={CoRL},
  year={2020},
}

@inproceedings{simeonov2020skills,
  title={A long horizon planning framework for manipulating rigid pointcloud objects},
  author={Simeonov, Anthony and Du, Yilun and Kim, Beomjoon and Hogan, Francois R. and Tenenbaum, Joshua and Agrawal, Pulkit and Rodriguez, Alberto},
  booktitle={CoRL},
  year={2020},
}

@article{neumann2014skills,
  title={Learning modular policies for robotics},
  author={Neumann, Gerhard and Daniel, Christian and Paraschos, Alexandros and Kupcsik, Andras and Peters, Jan},
  journal={Frontiers in Computational Neuroscience},
  year={2014},
}

@inproceedings{lee2020skills,
  title={Learning to coordinate manipulation skills via skill behavior diversification},
  author={Lee, Youngwoon and Yang, Jingyun and Lim, Joseph J},
  booktitle={ICLR},
  year={2020}
}

@inproceedings{jain2020actions,
  title={Generalization to New Actions in Reinforcement Learning},
  author={Ayush Jain and Andrew Szot and Joseph J. Lim},
  booktitle={ICML},
  year={2020},
}

@inproceedings{do2018affordance,
  title={AffordanceNet: An End-to-End Deep Learning Approach for Object Affordance Detection},
  author={Thanh-Toan Do and Anh Nguyen and Ian Reid},
  booktitle={ICRA},
  year={2018},
}

@article{kokic2020learning,
  title={Learning Task-Oriented Grasping From Human Activity Datasets},
  author={Mia Kokic and D. Kragic and Jeannette Bohg},
  journal={IEEE Robotics and Automation Letters},
  year={2020},
}

@inproceedings{xu2021daf,
  title={Deep Affordance Foresight: Planning Through What Can Be Done in the Future},
  author={Danfei Xu and Ajay Mandlekar and Roberto Martín-Martín and Yuke Zhu and Silvio Savarese and Li Fei-Fei},
  booktitle={ICRA},
  year={2021},
}

@inproceedings{mo2021where2act,
  title={Where2Act: From Pixels to Actions for Articulated 3D Objects},
  author={Kaichun Mo and Leonidas Guibas and Mustafa Mukadam and Abhinav Gupta and Shubham Tulsiani},
  booktitle={ICCV},
  year={2021},
}

@misc{klissarov2017learning,
      title={Learning Options End-to-End for Continuous Action Tasks}, 
      author={Martin Klissarov and Pierre-Luc Bacon and Jean Harb and Doina Precup},
      year={2017},
      eprint={1712.00004},
      archivePrefix={arXiv},
      primaryClass={cs.LG}
}

@inproceedings{tremblay2018dope,
 author = {Jonathan Tremblay and Thang To and Balakumar Sundaralingam and Yu Xiang and Dieter Fox and Stan Birchfield},
 title = {Deep Object Pose Estimation for Semantic Robotic Grasping of Household Objects},
 booktitle = {CoRL},
 year = 2018
}

@inproceedings{dalal2021raps,
  title={Accelerating Robotic Reinforcement Learning via Parameterized Action Primitives}, 
  author={Murtaza Dalal and Deepak Pathak and Ruslan Salakhutdinov},
  year={2021},
  booktitle={NeurIPS},
}

\clearpage
\appendix
\section{Implementation Details}\label{appsec:impl}

\subsection{Behavior Primitives}\label{appsec:primitives}
We elaborate on the manipulation primitives that we outlined in \cref{sec:primitives}: reaching, grasping, pushing, gripper release, and atomic.
We classify all primitives that are not the atomic primitive as \textit{non-atomic} primitives.
Under the hood, all non-atomic primitives execute a variable sequence of atomic actions, either until the primitive is successfully executed or until a time limit is reached.
All atomic actions specifically interface with the Operational Space Control (OSC) controller, which has 5 degrees of freedom: 3 degrees to control the position of the end effector, 1 degree to control the yaw angle, and (for all tasks but wiping) 1 degree to open and close the gripper.

We elaborate further on our non-atomic primitives.
The gripper release primitive executes a fixed number of atomic actions to open the gripper.
The reaching, grasping, and pushing primitives are hard-coded closed-loop controllers that all entail a reaching phase, either for reaching the starting location (for pushing) or for reaching the final location (for reaching and grasping).
To implement this functionality for table-top environments (all except door), the robot first lifts its end effector to a pre-specified height, then hovers to the target XY position, and finally lowers its end effector to the target location.
For other environments (door), the robot moves to toward the target location directory via the OSC controller.
During this reaching phase, the reaching primitive keeps its gripper closed (except for the non-tabletop environments like door) and the grasping and pushing primitives keep their grippers open.
The grasping and pushing primitives can be configured to achieve a specified yaw angle, which they satisfy during the reaching phase simultaneously while applying end effector displacements.
Upon reaching, the grasping primitive emulates grasping by closing its gripper and the pushing the primitive emulates pushing by applying end effector displacements in a specified direction.

\subsection{Algorithm}\label{appsec:algo}
Our algorithm implementation is based on Soft Actor-Critic.
Our algorithm alternates between collecting on-policy transitions in the environment and performing off-policy training on data sampled from the replay buffer.
Training specifically entails optimizing the Q network, task policy, and parameter policy via gradient descent.
As in the original SAC implementation, we use the reparameterization trick with respect to the parameter policy loss in order to reduce the variance of our gradient estimates.
While we assume continuous primitive parameters we can also represent discrete parameters and apply reparameterization with the Gumbel-Softmax trick~\citep{jang2017categorical,maddison2017concrete}.
We provide a full outline of our algorithm in \Cref{algo:ours}. Code is publicly available at \url{https://github.com/UT-Austin-RPL/maple}.

\begin{algorithm*}
\caption{\METHODLONGHIGHLIGHT\, (\METHODNOSPC)}
\label{algo:ours}
\begin{algorithmic}[1]
\STATE Initialize Q network $\Q(s, a, x)$, task policy $\pihip(a | s)$, parameter policy $\pilop(x | s, a)$, replay buffer $\mathcal{D}$ 
\FOR{iteration $1, \dots, N$}
    \FOR{episode $1, \dots, M$} \COMMENT{Exploration Phase}
    \STATE Initialize timer $t \leftarrow 0$
    \STATE Initialize episode $s_0$
    \WHILE{episode not terminated}
        \STATE Sample primitive type $a_t$ from task policy $\pihip(a_t | s_t)$
        \STATE Sample primitive parameters $x_t$ from parameter policy $\pilop(x_t | s_t, a_t)$
        \STATE Truncate sampled parameters to dimension of sampled primitive $x_t \leftarrow x_t[:d_{a_t}]$
        \STATE Execute $a_t$ and $x_t$ in environment, obtain reward $r_t$ and next state $s_{t+1}$
        \STATE Add affordance score to reward $r_t \leftarrow r_t + \lambda s_{\text{aff}}(s_t, x_t; a_t)$
        \STATE Add transition to replay buffer $\mathcal{D} \leftarrow \mathcal{D} \cup \{ s_t, a_t, x_t, r_t, s_{t+1}\}$
        \STATE Update timer $t \leftarrow t+1$
    \ENDWHILE
    \ENDFOR
    \FOR{training step $1, \dots,  K$} \COMMENT{Training Phase}
        \STATE Update Q network: $\theta \leftarrow \theta - \lambda_{lr} \nabla_\theta J_Q(\theta)$
        \STATE Update task policy: $\phi \leftarrow \phi - \lambda_{lr} \nabla_\phi J_{\pihi}(\phi)$
        \STATE Update parameter policy: $\psi \leftarrow \psi - \lambda_{lr} \nabla_\psi J_{\pilo}(\psi)$
    \ENDFOR
\ENDFOR
\end{algorithmic}
\end{algorithm*}

\subsection{Affordance Score}\label{appsec:affordance}
We elaborate on the affordance score introduced in \Cref{sec:affordance}:
\begin{align*}
s_{\text{aff}}(s, x; a) &= \max_{p \in P} \ 1 - \tanh\Big(\max(\Vert x_{reach} - p \Vert - \tau, 0)\Big)
\end{align*}
The keypoint $p$ is dependent on the primitive $a$ and the current state $s$.
For example for the cleanup task, the keypoint for the pushing primitive is the location of a pushable object (the jello box), the keypoint for a grasping primitive is the location of a graspable object (the spam can), and the keypoint for the reaching primitive is the location of the bin.
If there are multiple keypoints of interest we calculate the affordance score corresponding to each keypoint and consider the maximum score.
If no applicable keypoint exists for a primitive (e.g.\ there are no pushable objects in door opening) we give an affordance score of $0$.
By default we set the threshold $\tau$ to $0.03$ for grasping, $0.06$ for reaching, and $0.12$ for pushing.
There are a few exceptions for tasks that need larger affordance regions for reaching large objects.

\subsection{Flat Baseline}\label{appsec:flat}
We considered two variants for our flat baseline.
One variant, which has been explored by prior work~\citep{xiali2020relmogen,hausknecht2016deep}, outputs a distribution over the primitive type $a$ and the parameters $\{x^1, x^2, \cdots, x^k\}$ for all primitives.
As we discussed in \Cref{sec:algo} under this approach the number of policy outputs scales linearly with the total number of primitive parameters, which can lead to optimization difficulties for large behavior libraries. 
Empirically we found this to be the case, as we were unable to make any progress on any of our tasks despite extensive hyperparamter tuning.
Neunert et al.~\citep{neunert2020continuousdiscrete} proposed an alternative approach of replacing the distribution over all parameter outputs with the ``one size fits all'' distribution that we described in \Cref{sec:algo}.
Parameter selection occurs by \textit{independently} sampling a primitive type $a$ and parameters $x$, and subsequently truncating the sampled parameters by the dimension of the sampled primitive type.
This sampling strategy was also adopted by Lee et al.~\cite{lee2020skills}.
We note that this independent sampling process is in contrast to our two-stage hierarchical process.
We adopted this variant as our flat baseline, and in \Cref{plot:main-results} we see that it often leads to sub-optimal performance.
We hypothesize that this is due the fact that the parameter selection process is not informed by the primitive type selection process,
which reduces the agent's utility especially when dealing with primitives that have heterogeneous parameter structures.

\subsection{Open Loop Baseline}\label{appsec:openloop}
Our open loop baseline follows an open loop task schema, and is inspired from Chitnis et al.~\citep{chitnis2020schema}.
The open loop baseline and our method share identical implementations, except for the input to the task policy: our method takes in the current environment observation while the open loop baseline takes in only the current episode timestep.
We highlight that while our implementation is inspired from Chitnis et al.~\citep{chitnis2020schema}, there are notable differences.
Their update rule for the ``task policy'' (or equivalent thereof) does not use gradient descent, relies on on-policy sampling, and is designed for the sparse reward setting only.
We found these assumptions to be restrictive for our algorithmic and task setup, and we instead use gradient-based, off-policy reinforcement learning methods which can work with sparse or dense rewards.
Despite these differences, we believe that our open loop baseline captures the essence of the ideas proposed in Chitnis et al.~\citep{chitnis2020schema} -- namely that open loop task shemas can enable more efficient and effective learning.
As we show in \Cref{plot:main-results} however, we did not find this to be the case for the relatively more complex tasks in our suite of manipulation domains.

\subsection{DAC Baseline}\label{appsec:dac}
We considered a number of potential methods as our representative option-learning baseline.
While prominent prior work~\citep{bacon2016optioncritic,klissarov2017learning,zhang2019dac} has focused on learning options, we verified that Double Actor-Critic (DAC) achieves superior performance on the OpenAI HalfCheetah-v2 task and our lift task, so we chose DAC as our representative options baseline. We also considered Deep Skill Chaining~\citep{bagaria2020dsc}, another recent work that learns options; however it was not applicable to our manipulation domains given that it is designed primarily for goal-based navigation agents.
We used the implementation publicly released by the DAC authors \footnote{\url{https://github.com/ShangtongZhang/DeepRL/tree/DAC}}, and we adopted the hyperparameters that they suggested in their paper.

\section{Experimental Setup}\label{appsec:setup}

\subsection{Environments}\label{appsec:envs}
We conduct experiments on eight manipulation tasks of varying complexities, spanning diverse prehensile and non-prehensile behaviors. The first six come from the standard robosuite benchmark~\citep{robosuite2020}. We designed the last two (cleanup, peg insertion) to evaluate our method in multi-stage, contact-rich tasks.
We elaborate on each as follows:
\begin{figure}[H]
    \centering
    \begin{subfigure}[t]{0.22\linewidth}
        \includegraphics[width=\textwidth]{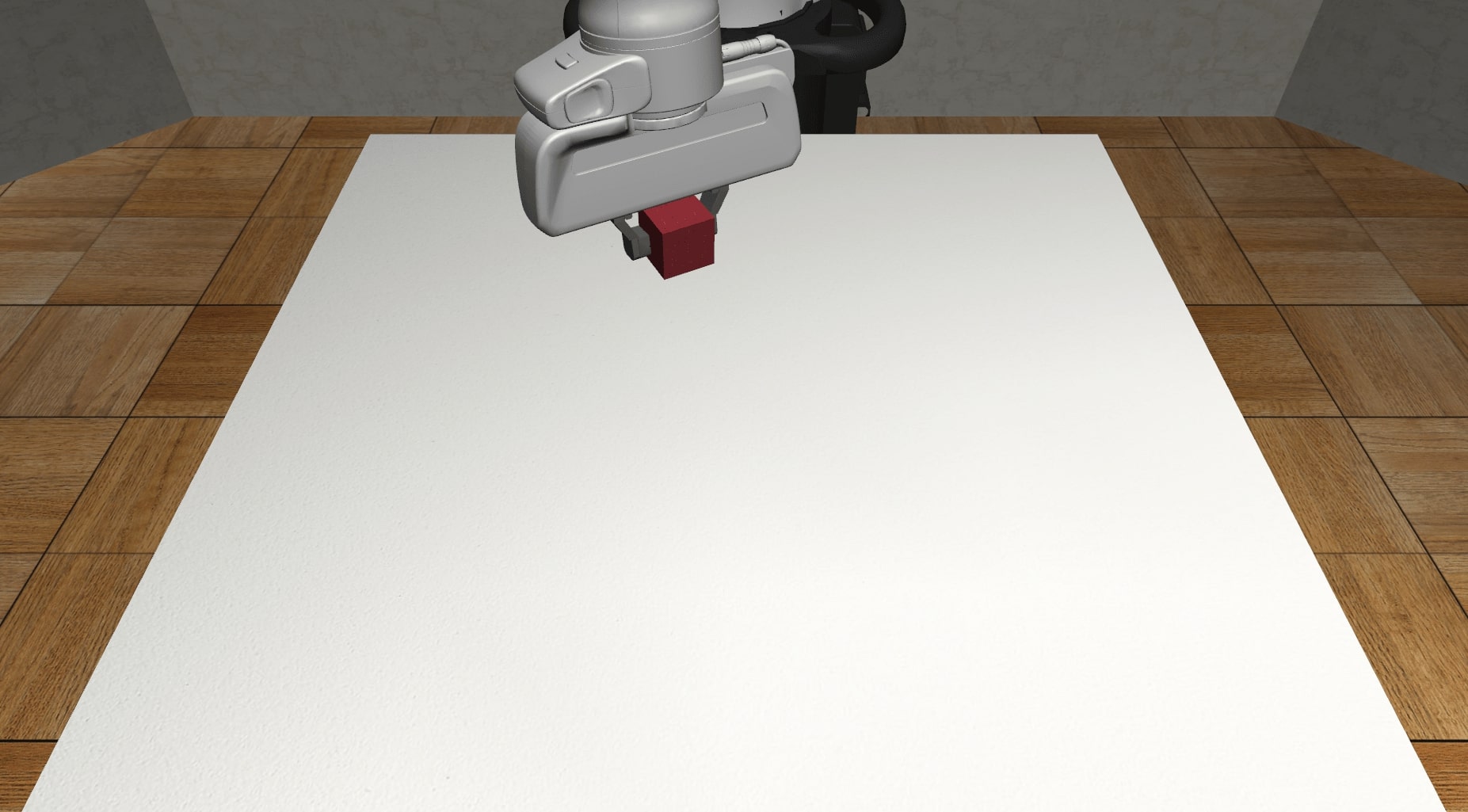}
        \caption*{\textbf{Lift:} the robot must pick up a cube and lift it above the table.}
    \end{subfigure}
    \hfill
    \begin{subfigure}[t]{0.22\linewidth}
        \includegraphics[width=\textwidth]{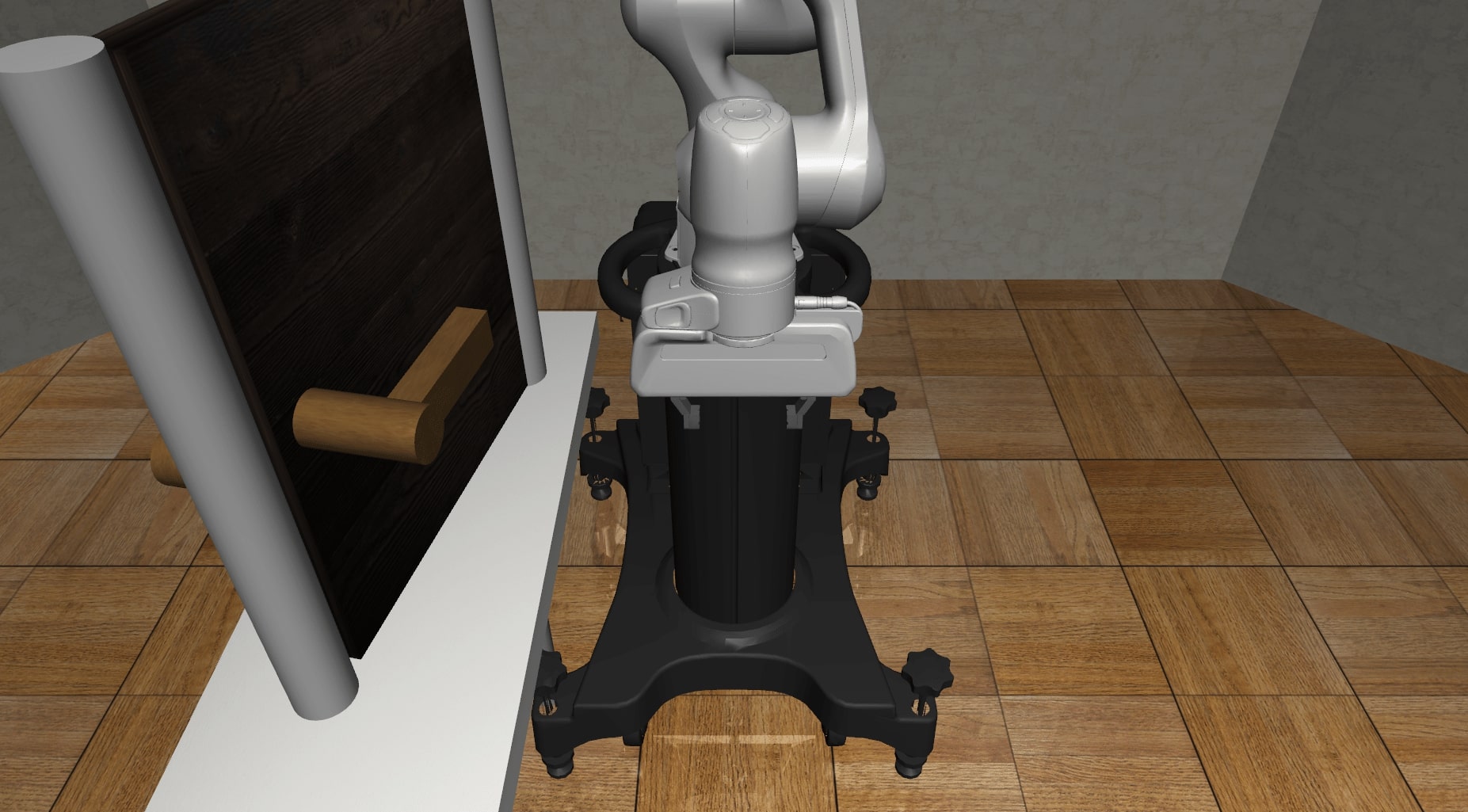}
        \caption*{\textbf{Door Opening:}
        the robot must turn the door handle and open the door.}
    \end{subfigure}
    \hfill
    \begin{subfigure}[t]{0.22\linewidth}
        \includegraphics[width=\textwidth]{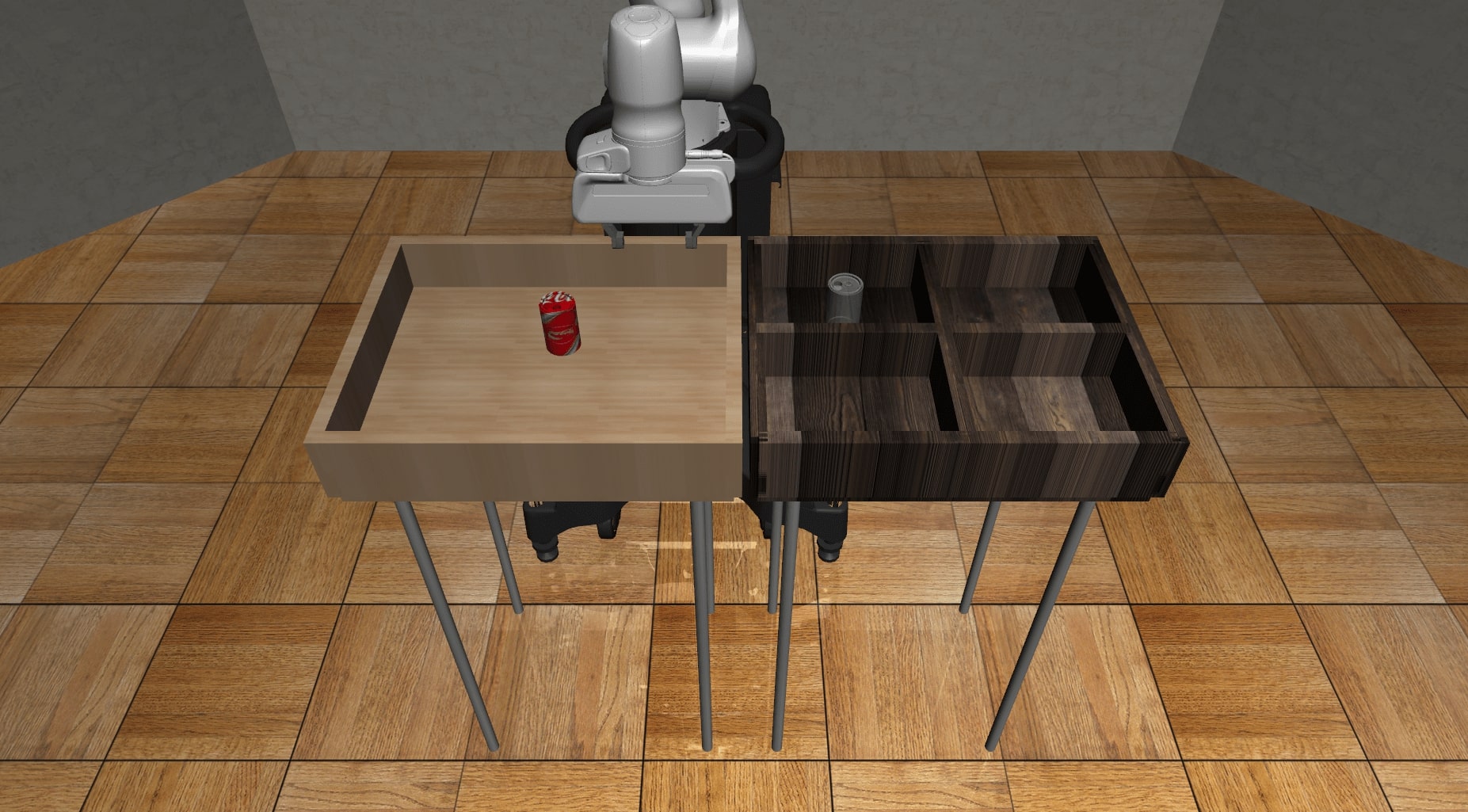}
        \caption*{\textbf{Pick and Place:}
        the robot must pick up a soda can and place it into a specific target compartment.}
    \end{subfigure}
    \hfill
    \vspace{20pt}
    \begin{subfigure}[t]{0.22\linewidth}
        \includegraphics[width=\textwidth]{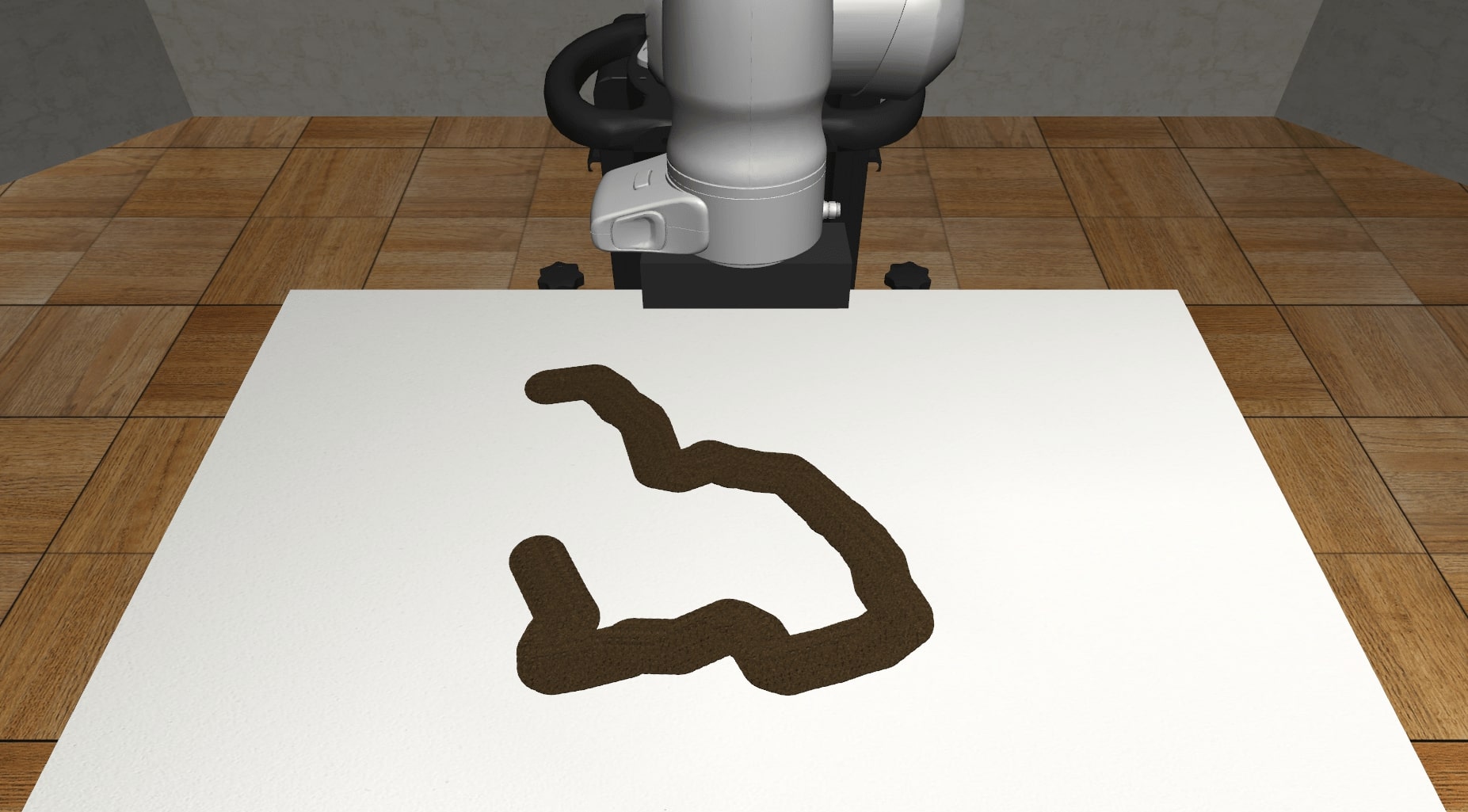}
        \caption*{\textbf{Wipe:}
        the robot must wipe a table containing spilled debris. A penalty is given if the robot presses too hard into the table.}
    \end{subfigure}
    \begin{subfigure}[t]{0.22\linewidth}
        \includegraphics[width=\textwidth]{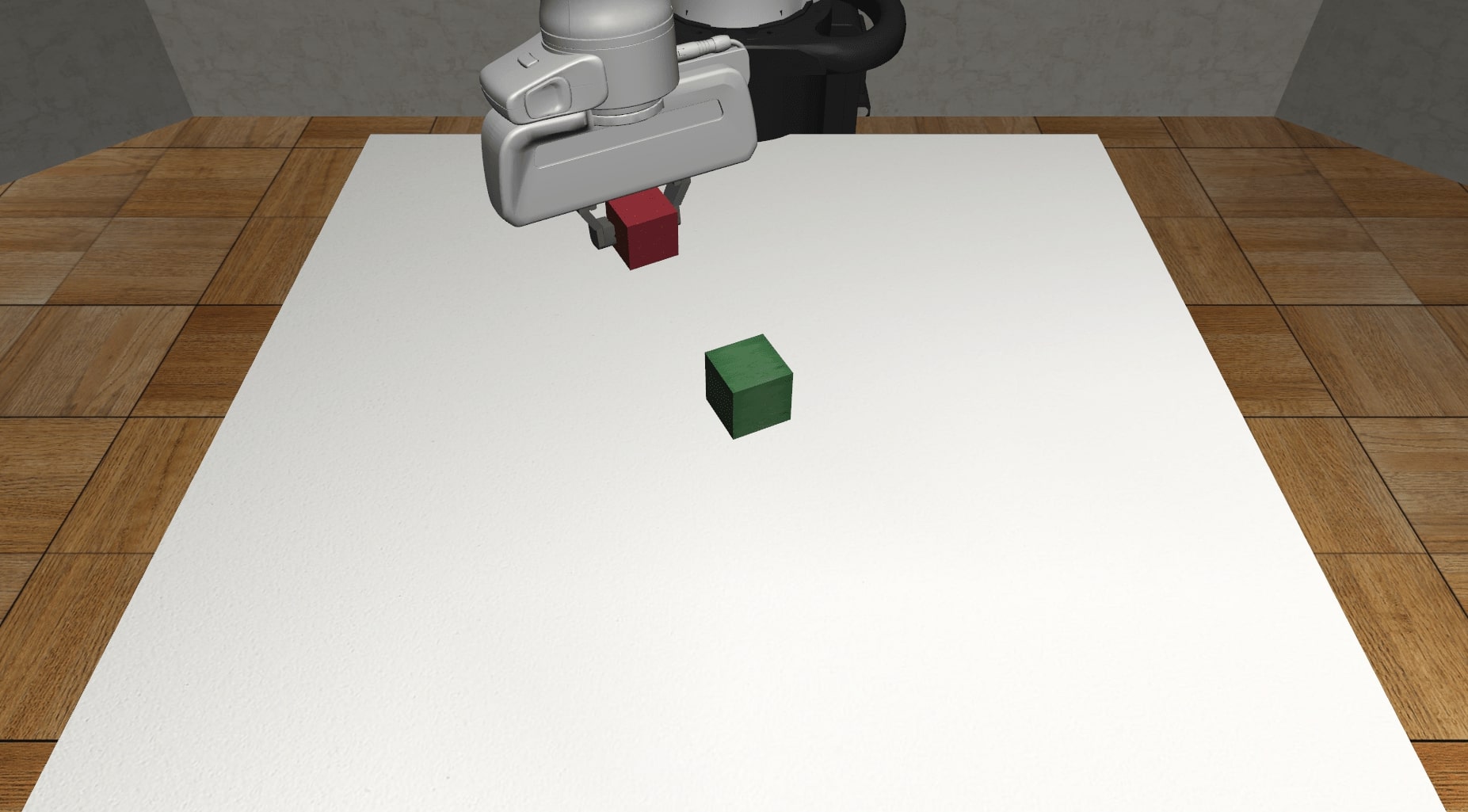}
        \caption*{\textbf{Stack:} the robot must stack a cube on top of another cube.}
    \end{subfigure}
    \hfill
    \begin{subfigure}[t]{0.22\linewidth}
        \includegraphics[width=\textwidth]{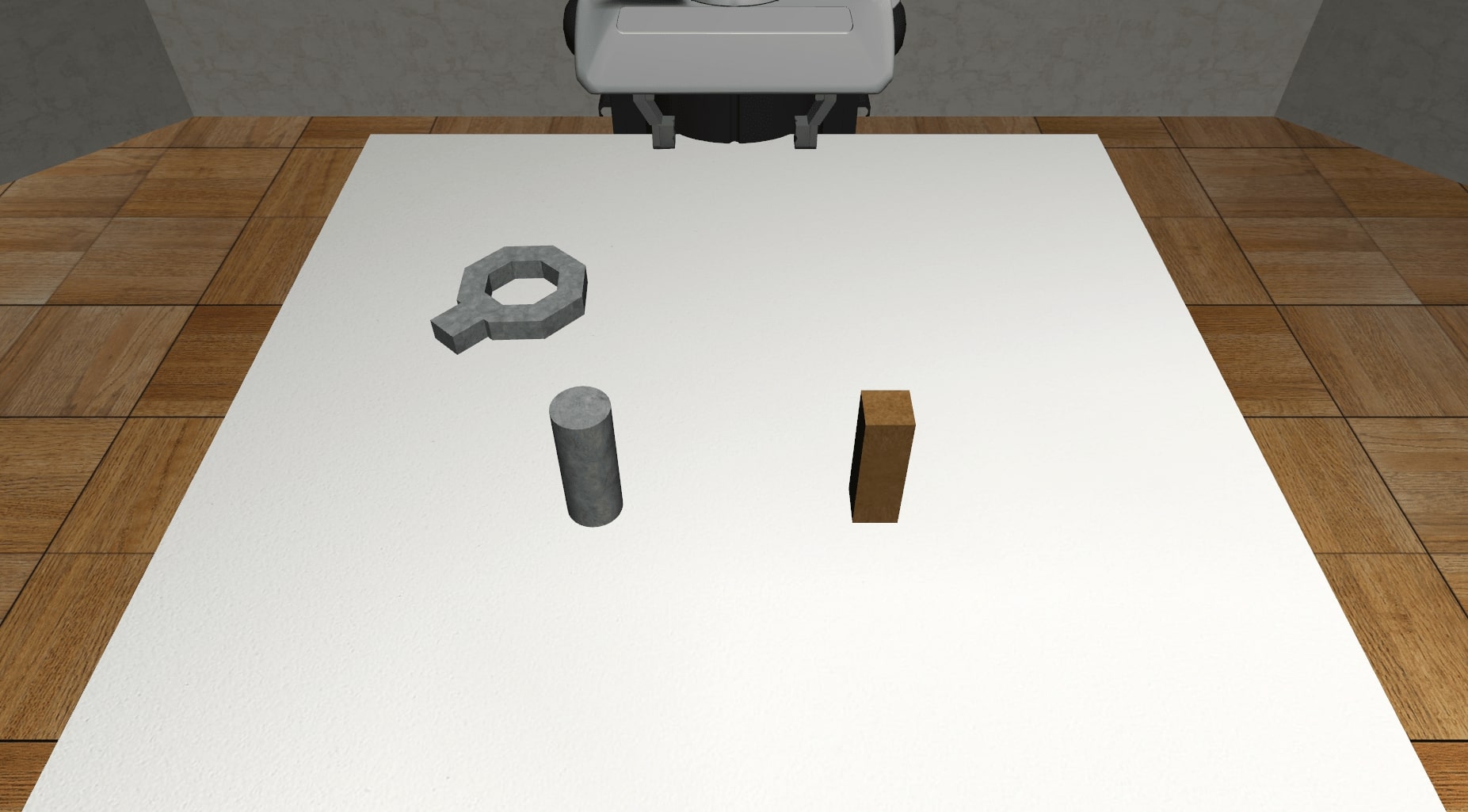}
        \caption*{\textbf{Nut Assembly:}
        the robot must fit a nut tool onto the round peg.}
    \end{subfigure}
    \hfill
    \begin{subfigure}[t]{0.22\linewidth}
        \includegraphics[width=\textwidth]{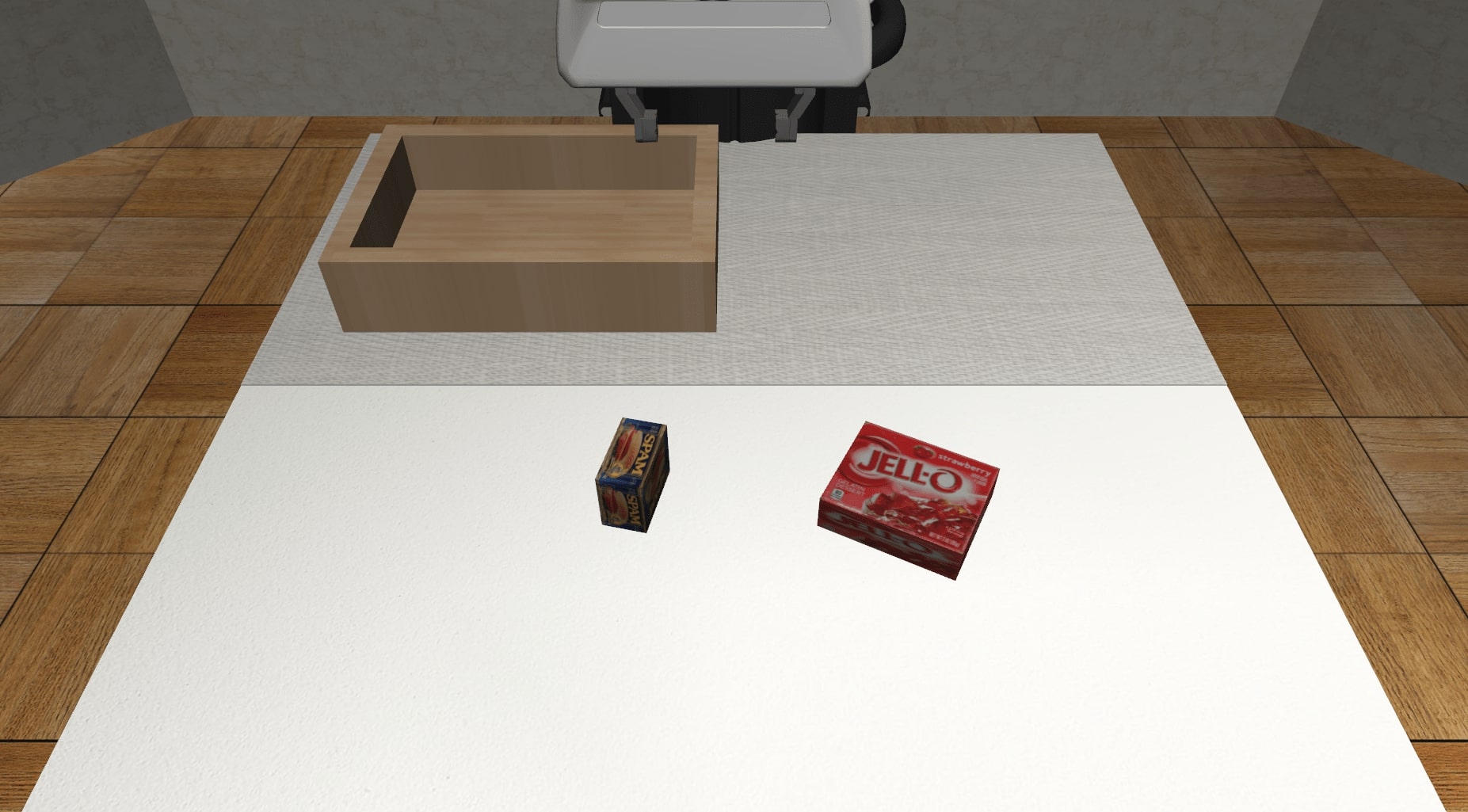}
        \caption*{\textbf{Cleanup:}
        the robot must store a spam can into a storage bin and store a jello box at the upper right corner.}
    \end{subfigure}
    \hfill
    \begin{subfigure}[t]{0.22\linewidth}
        \includegraphics[width=\textwidth]{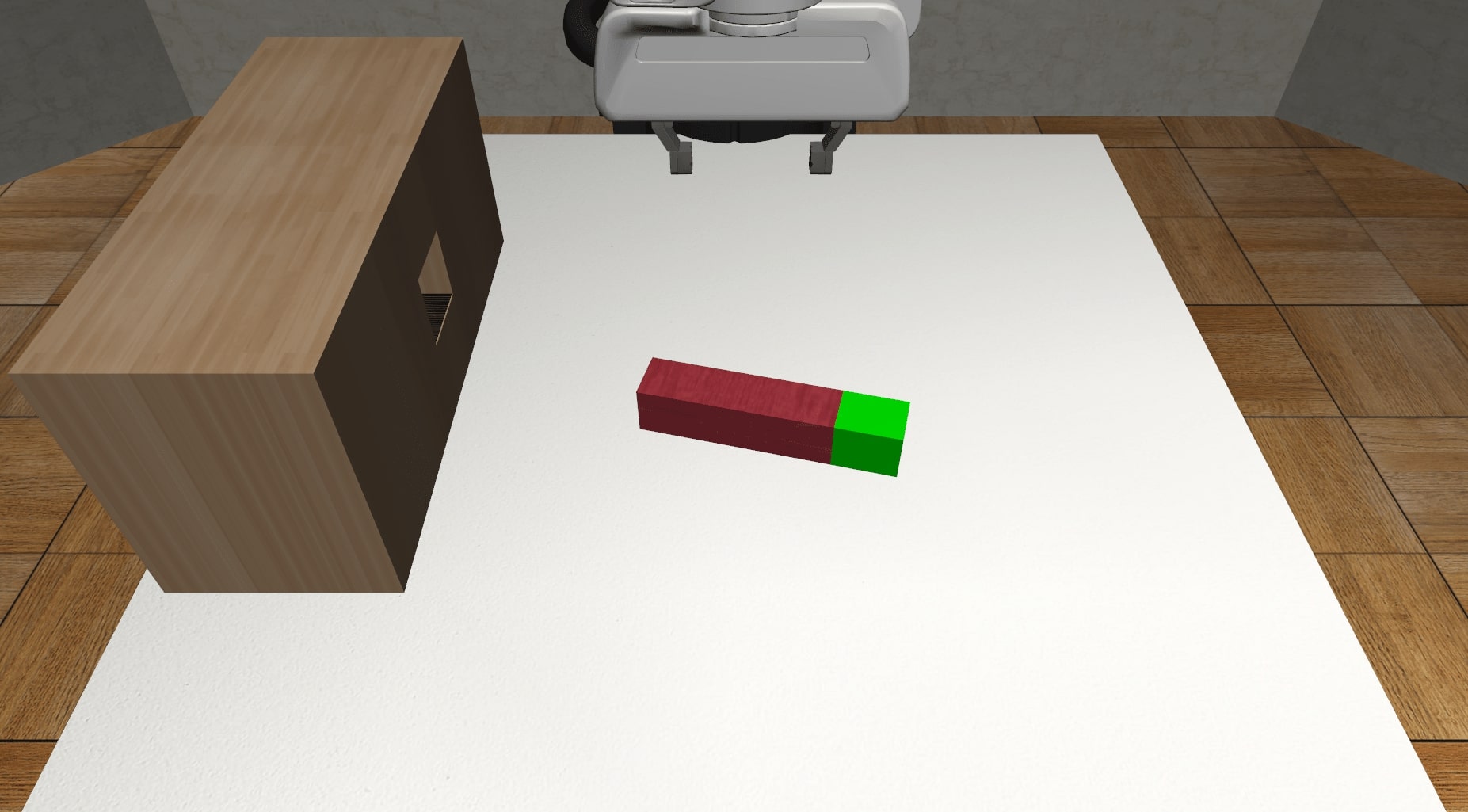}
        \caption*{\textbf{Peg Insertion:}
        the robot must pick up the peg and insert it into the opening of a wooden block.}
    \end{subfigure}
    \centering
\end{figure}


\subsection{Training}
We provide a full list of our algorithm hyperparameters in \Cref{table:hyperparams}.
We note a few additional details.
For a consistent comparison across baselines, our episode lengths are fixed to $150$ \textit{atomic} timesteps, meaning that we execute a variable number of primitives until we have exceeded the maximum number of atomic actions for the episode.
Also, for the first 600k environment steps we set the target entropy for the task policy and parameter policy to a high value to encourage higher exploration during the initial stages of training.
\begin{table*}
\vspace{15pt}
\centering
\caption{Hyperparameters for our algorithm}
\begin{tabular}{c|c}
\hline
\textbf{Hyperparameter} & \textbf{Value}\\
\hline
Hidden sizes (all networks) & $256, 256$\\
Q network and policy activation & ReLU\\
Q network output activation & None\\
Policy network output activation & tanh\\
& \\
Optimizer & Adam\\
Batch Size & $1024$\\
Learning rate (all networks) & $3\mathrm{e}{-5}$\\
Target network update rate $\tau$ & $1\mathrm{e}{-3}$\\
& \\
\# Training steps per epoch & $1000$\\
\# (Low-level) exploration actions per epoch & $3000$\\
Replay buffer size & $1\mathrm{e}{6}$\\
Episode length (\# low-level actions) & 150 (except wipe, 300)\\
& \\
Discount factor & $0.99$\\
Reward scale & $5.0$\\
Affordance score scale $\lambda$ & $3.0$\\
Automatic entropy tuning & True\\
Target Task Policy Entropy & $0.50 \times log(k)$, $k$ is number of primitives \\
Target Parameter Policy Entropy & $-\max_a d_a$\\
\hline
\end{tabular}
\label{table:hyperparams}
\end{table*}

\subsection{Evaluation}

We elaborate on the evaluation protocol for our experiments in \Cref{plot:main-results}.
We evaluate each experimental variant (combination of task and method) over 5 seeds and we (1) plot the agent's rewards throughout training and (2) report the task success rate at the end of training.
Specifically for the reward plots, we evaluate the agent's average episodic task rewards (excluding the affordance reward) at regularly spaced training checkpoints every $30k$ environment exploration steps.
The episodic rewards are averaged over 20 episodes and are normalized between $0$ and $100$, where $100$ corresponds to the agent receiving the maximum possible reward at every single timestep of the episode.
We post-process the plots, showing the moving average of results over the last $150k$ environment steps.
For reporting the final task success rate, we load the final training checkpoint and report the average task success rate over 20 episodes. Success rates for our tasks are defined as follows:
\begin{itemize}
    \item Lift: whether the block is above a height threshold
    \item Door: whether the door angle is past a threshold
    \item Pick and Place: whether the can is in the correct target bin and the robot is not holding the can
    \item Wipe: whether all of the debris is wiped off the table
    \item Stack: whether the smaller cube is on top of the larger cube and the robot is not holding either cube
    \item Nut Assembly: whether the nut is fitted completely onto the round peg and the robot is not holding the nut
    \item Cleanup: whether the spam can is in the bin and the jello box is within a threshold distance away from the table corner
    \item Peg Insertion: whether the peg is inserted into wooden block past a threshold distance
\end{itemize}
Final task success rates for all baselines across all tasks are outlined in \Cref{table:success}.
\begin{table*}
    \vspace{25pt}
    \centering
    \caption{Final Task Success Rates (\%)}
    \label{table:success}
    \begingroup
    \fontsize{7.5pt}{13pt}\selectfont
    \setlength\tabcolsep{0.6pt}
    
    \newcolumntype{E}{>{\arraybackslash} m{2.2cm} }
    \newcolumntype{F}{>{\centering\arraybackslash} m{1.4cm} }
    
    \begin{tabular}{ |E|F|F|F|F|F|F|F|F| } 
        \hline
        & \textbf{Lift}
        & \textbf{Door}
        & \textbf{Pick and Place} &
        \textbf{Wipe} & \textbf{Stack} & \textbf{Nut Assembly} & \textbf{Cleanup} & \textbf{Peg Insertion} \\
        \hline
        Atomic~\cite{haarnoja2018sac}
        & \textbf{98.0 $\pm$ 2.4}
        & 0.0 $\pm$ 0.0
        & 0.0 $\pm$ 0.0 &
        18.0 $\pm$ 18.3 &
        38.0 $\pm$ 28.7 &
        0.0 $\pm$ 0.0 &
        0.0 $\pm$ 0.0 &
        0.0 $\pm$ 0.0 \\
        Flat~\cite{lee2020skills,neunert2020continuousdiscrete}
        & 61.0 $\pm$ 47.8
        & \textbf{100.0 $\pm$ 0.0}
        & 1.0 $\pm$ 2.0 &
        22.0 $\pm$ 9.8 &
        \textbf{98.0 $\pm$ 2.4} &
        0.0 $\pm$ 0.0 &
        0.0 $\pm$ 0.0 &
        8.0 $\pm$ 13.6 \\
        Open Loop~\cite{chitnis2020schema}
        & 43.0 $\pm$ 43.5
        & \textbf{100.0 $\pm$ 0.0}
        & 81.0 $\pm$ 38.0 &
        16.0 $\pm$ 4.9 &
        85.0 $\pm$ 3.2 &
        0.0 $\pm$ 0.0 &
        0.0 $\pm$ 0.0 &
        0.0 $\pm$ 0.0 \\
        DAC~\cite{zhang2019dac}
        & 75.0 $\pm$ 12.7
        & 0.0 $\pm$ 0.0
        & 0.0 $\pm$ 0.0 &
        0.0 $\pm$ 0.0 &
        0.0 $\pm$ 0.0 &
        0.0 $\pm$ 0.0 &
        0.0 $\pm$ 0.0 &
        16.0 $\pm$ 32.0 \\
        \METHOD (Non-At.)
        & \textbf{100.0 $\pm$ 0.0}
        & \textbf{100.0 $\pm$ 0.0}
        & \textbf{100.0 $\pm$ 0.0} &
        \textbf{42.0 $\pm$ 9.8} &
        \textbf{99.0 $\pm$ 2.0} &
        93.0 $\pm$ 14.0 &
        \textbf{100.0 $\pm$ 0.0} & 0.0 $\pm$ 0.0 \\
        \METHOD (ours)
        & \textbf{100.0 $\pm$ 0.0}
        & \textbf{100.0 $\pm$ 0.0}
        & \textbf{95.0 $\pm$ 7.7} &
        \textbf{42.0 $\pm$ 11.7} &
        \textbf{98.0 $\pm$ 2.4} &
        \textbf{99.0 $\pm$ 2.0} &
        91.0 $\pm$ 5.8 &
        \textbf{100.0 $\pm$ 0.0} \\
        \hline
    \end{tabular}
    \endgroup
\end{table*}

We also elaborate on the evaluation protocol for the compositionality scores that we report in \Cref{fig:task-sketches}.
Our compositionality scores are averaged over 5 seeds for each task.
For each seed we sample 50 task sketches from the last training checkpoint and we discard sketches that did not correspond to the agent solving the task.
Of these remaining task sketches we calculate the compositionality score according to \Cref{eq:comp-metric}.

\subsection{Task Sketch Transfer Experiments}
For our task sketch experiments we first extract a set of task sketches $\{K_1, \cdots, K_n\}$ from the source task.
We subsequently select the sketch $K$ that has the lowest Levenshtein distance with all other task sketches.
In the case of our pick and place task this was \{Grasp, Reach, Release\}.
Once we have extracted the sketch $K$ from the source task, we train on the target task with a fixed task sketch of $K$.
For each episode we iterate through the sequence of primitives in $K$, repeating each primitive up to 5 times until the agent receives high affordance reward, before moving onto the next primitive in the sketch.
Upon executing all of the primitives in the task sketch the agent executes 10 atomic primitives to satisfy any behaviors that it was not able to fulfill with the sketch alone, and then the episode terminates.

\subsection{Real-World Experiments}
We performed evaluations on two real-world manipulation tasks:
\begin{figure}[H]
    \centering
    \begin{subfigure}[t]{0.47\linewidth}
        \includegraphics[width=\textwidth]{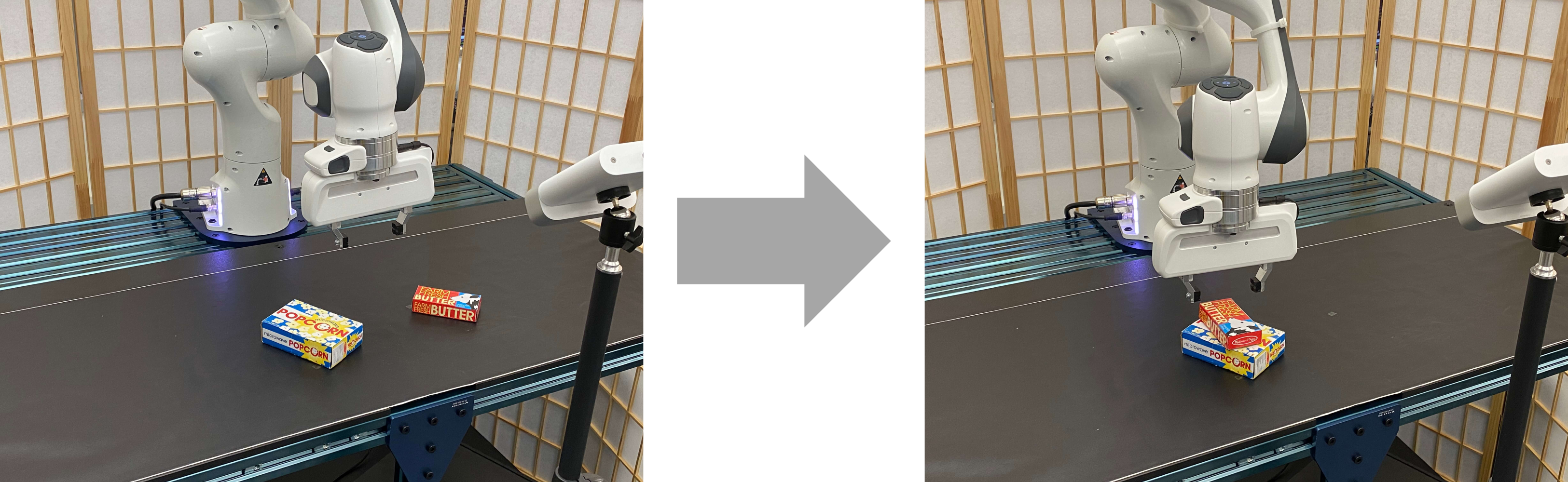}
        \caption*{\textbf{Stack:} the robot must pick up the butter box and stack it on top of the popcorn box.}
    \end{subfigure}
    \hfill
    \begin{subfigure}[t]{0.47\linewidth}
        \includegraphics[width=\textwidth]{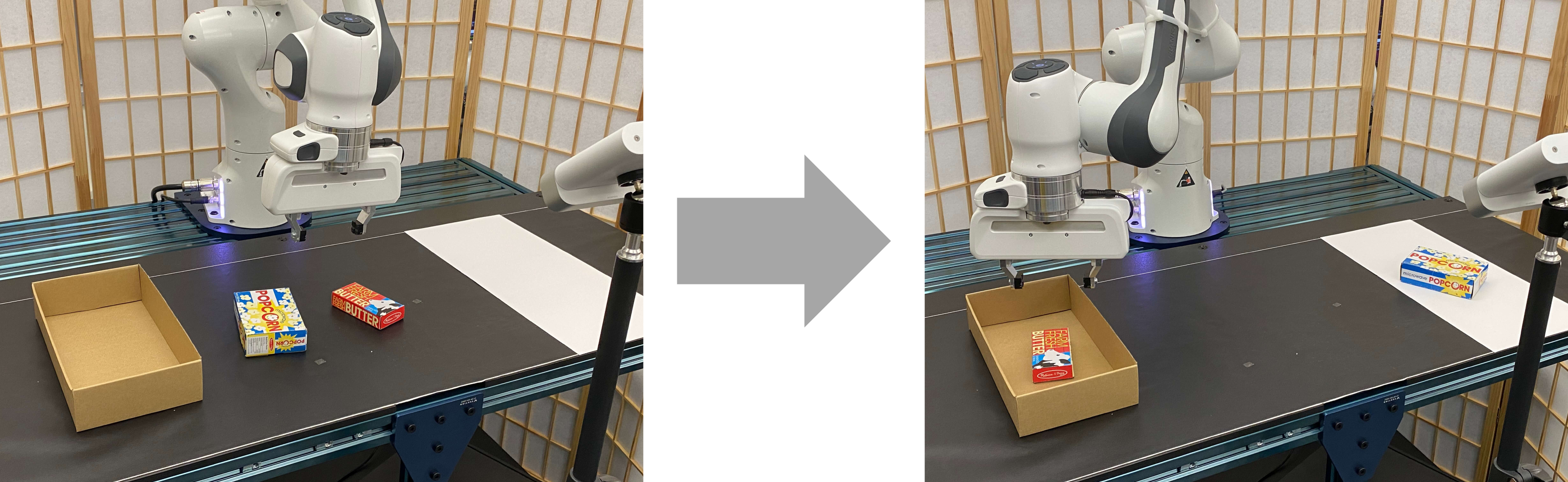}
        \caption*{\textbf{Cleanup:} the robot must (1) pick up the butter box and place it into the bin and (2) push the popcorn to the right side of the table (the white area).}
    \end{subfigure}
    \centering
\end{figure}
Both tasks resemble the simulated stack and cleanup tasks outlined in \Cref{appsec:envs}, but have differences in the size of the objects, table size, and workspace layout.
To account for these differences we designed variations of our existing simulated stack and cleanup tasks to match the characteristics of our real-world tasks.
We trained policies in simulation (until convergence) and transferred them to the real-world for evaluation.

For perception we use the robot proprioception data from the robot's on-board sensors, in conjunction with pose estimates of the objects using the deep object pose estimation system~\citep{tremblay2018dope} paired with a single Microsoft Kinect camera.
Under this setup the objects are sometimes out of the camera view or are occluded by the robot arm; in these cases the pose estimator does not return estimates of the object.
We mostly alleviate such conditions through a three step procedure: (1) the robot lifts its end-effector to a pre-determined location in the air where occlusions are minimized, (2) the pose estimator re-computes the poses of the objects, (3) the robot moves back to its initial location.
At the end of step (3), the pose of objects are assumed to be the pose estimates from step (2), with the exception of objects that were moved by the robot during step (1).
Such objects comprise objects that the robot was already holding before step (1) and that the robot subsequently lifted into the air during step (1).
For such objects, we compute the pose of the object as the final robot pose in addition to the relative pose difference of the robot end effector and object during step (2).
In addition to handling occlusions, we noticed that the pose estimation system routinely made small errors when estimating the position and orientation of the objects.
To minimize the influence of such errors, we hard-coded the pitch and yaw angles of the objects (as they were always flat either on the table or in the air) and the height of the object whenever it was detected to lie on the table.
These constraints were necessary to ensure reliable perception estimates, but we anticipate that with improved perception systems in the future such constraints can be relaxed.

For evaluation, we performed 30 trials for each task, where the robot was allowed a maximum of 20 primitive calls per episode.
We recorded an average success rate of $93\%$ for stack (in contrast to $97\%$ in simulation), and $83\%$ for cleanup (in contrast to $93\%$ in simulation).
Most failures were either due to the robot repeatedly applying poor grasping actions or the robot hitting its joint limits, subsequently triggering a safety call to halt the robot.

\end{document}